\documentclass{article}

\usepackage{microtype}
\usepackage{graphicx}
\usepackage{subfigure}
\usepackage{booktabs} 
\usepackage{amsmath}
\usepackage{amssymb}
\usepackage{algorithm}
\usepackage{algorithmic}
\usepackage{multirow}

\usepackage{subcaption}
\usepackage{mathtools}
\usepackage{amsthm}
\usepackage{natbib}
\usepackage{fancyvrb}
\usepackage{tikz}
\usetikzlibrary{arrows.meta, positioning}
\usepackage[most]{tcolorbox}
\usepackage{float}
\usepackage{xcolor}
\usepackage{placeins}
\usepackage{bm}
\usepackage{amsthm}

\theoremstyle{definition}

\newcommand{\State}{\STATE}
\newcommand{\For}{\FOR}
\newcommand{\EndFor}{\ENDFOR}

\usepackage{hyperref}
\usepackage[capitalize,noabbrev]{cleveref}

\usepackage[accepted]{mlsys2025}

\mlsystitlerunning{RAMP: Reinforcement Adaptive Mixed-Precision Quantization for Efficient On-Device LLM Inference}

\begin{document}

\twocolumn[
\mlsystitle{RAMP: Reinforcement Adaptive Mixed-Precision Quantization for
Efficient On-Device LLM Inference}

\begin{mlsysauthorlist}
\mlsysauthor{Arpit Singh Gautam}{}
\mlsysauthor{Saurabh Jha}{}
\end{mlsysauthorlist}

\mlsyscorrespondingauthor{Arpit Singh Gautam}{arpitsinghgautam777@gmail.com}

\mlsyskeywords{Large language models, post-training quantization, mixed precision, reinforcement learning, on-device inference}
\vskip 0.3in

\begin{abstract}
Post-training quantization is essential for deploying large language models (LLMs) on resource-constrained hardware, yet state-of-the-art methods enforce uniform bit-widths across layers, yielding suboptimal accuracy-efficiency trade-offs. We present RAMP (Reinforcement Adaptive Mixed-Precision), an off-policy Soft Actor-Critic framework that learns per-layer bit-width assignments to minimize perplexity under a global bit budget. The policy conditions on an 11-dimensional embedding of activation statistics, weight properties, and structural descriptors, enabling zero-shot transfer across model families and scales. To enable stable sub-4-bit quantization, we introduce Scale Folding, a preconditioning technique that migrates activation outliers into weights via per-channel scaling and normalization-layer compensation. A quality-prioritized reward with asymmetric penalties and budget cliffs drives rapid convergence. On Llama-2-7B, RAMP achieves 5.54 perplexity at 3.68\,GB (3.65 effective bits), outperforming uniform 4-bit AWQ (5.60 at 3.90\,GB) and GPTQ by 6\% in size and 1--3\% in quality. Critically, a policy trained only on Llama-2-7B generalizes zero-shot to Llama-2-13B and Mistral-7B, often surpassing target-specific training, supporting the hypothesis that quantization sensitivity is primarily architectural. The HALO pipeline exports allocations to GGUF format for kernel-free inference on CPUs, GPUs, and edge devices, retaining 99.5\% of FP16 commonsense reasoning performance.
\end{abstract}
]

\section{Introduction}
\label{sec:introduction}

\subsection{The Memory Wall in Large Language Models}
\label{sec:intro_memory_wall}

The advent of large language models (LLMs) has fundamentally transformed natural language processing. Models such as GPT-4~\citep{openai2023gpt4}, Llama-2~\citep{touvron2023llama2}, Llama-3~\citep{meta2024llama3}, and Mistral~\citep{jiang2023mistral} achieve state-of-the-art performance across tasks including machine translation, code generation, and multi-step reasoning. These advances have driven widespread adoption in both research and commercial settings.

However, the scale of modern LLMs introduces a critical deployment bottleneck: the growing disparity between model memory requirements and available hardware capacity, commonly termed the \emph{memory wall}. For example, Llama-2-13B (13 billion parameters) requires approximately 26\,GB in FP16 format, exceeding the memory of many consumer GPUs. Even Llama-2-7B demands about 13.5\,GB in FP16, leaving limited headroom for activations during inference.

This constraint severely restricts deployment on:
\begin{itemize}
    \item Edge devices with constrained memory (mobile phones, IoT devices, embedded systems),
    \item Consumer-grade GPUs (e.g., RTX 3090, RTX 4090),
    \item Cost-sensitive cloud environments where memory bandwidth and capacity dominate inference costs,
    \item Privacy-sensitive applications that require on-device inference.
\end{itemize}

The economic and environmental implications are significant. High-end datacenter GPUs capable of hosting unquantized 13B-scale models cost tens of thousands of dollars, while consumer alternatives are substantially cheaper yet insufficient. Moreover, large-scale cloud-based inference contributes meaningfully to the carbon footprint of AI systems.

\Cref{tab:memory_wall} quantifies this memory wall for representative models relative to a typical consumer GPU memory limit.

\begin{table}[t]
\centering
\caption{Model memory footprint in FP16 and viability on 24\,GB consumer GPU}
\label{tab:memory_wall}
\resizebox{\columnwidth}{!}{
\begin{tabular}{lcccc}
\toprule
\textbf{Model} & \textbf{Parameters} & \textbf{FP16 Size} & \textbf{24\,GB GPU} & \textbf{Viable?} \\
\midrule
Llama-2-7B     & 7.0B  & 13.5\,GB & 24\,GB & Marginal \\
Llama-2-13B    & 13.0B & 26.0\,GB & 24\,GB & No \\
Llama-3-8B     & 8.0B  & 16.1\,GB & 24\,GB & No \\
Mistral-7B     & 7.0B  & 14.5\,GB & 24\,GB & Marginal \\
GPT-3.5        & 175B+ & 350+\,GB & 24\,GB & Impossible \\
\bottomrule
\end{tabular}
}
\end{table}

\subsection{Limitations of Existing Quantization Methods}
\label{sec:intro_limitations}

Post-training quantization (PTQ) is the primary technique for reducing LLM memory footprint. By representing weights and activations in lower bit-widths (typically 4--8 bits), PTQ achieves 4--8$\times$ compression with modest accuracy loss. Recent methods such as GPTQ~\citep{frantar2023gptq} and AWQ~\citep{lin2023awq} demonstrate that 4-bit quantization can preserve near-full-precision performance on many tasks.

Nevertheless, current approaches exhibit three important limitations.

\subsubsection{Uniform Bit-Width Allocation}
State-of-the-art PTQ methods apply a uniform bit-width across all layers. This ignores substantial variation in layer sensitivity to quantization noise. In Transformer-based architectures, embedding layers, attention output projections, and final language-modeling heads are particularly sensitive, as errors here propagate globally or directly affect predictions. In contrast, many intermediate MLP layers exhibit redundancy and greater tolerance to low-precision representations.

Uniform allocation therefore over-allocates bits to robust layers while under-allocating them to sensitive ones, resulting in a suboptimal accuracy--efficiency trade-off.

\subsubsection{Lack of Transferability Across Models}
Existing methods require costly per-model optimization and calibration. For example, GPTQ performs layer-wise Hessian-based optimization with $\mathcal{O}(d^2)$ complexity (where $d$ is the hidden dimension), which becomes prohibitive for $d \geq 4096$. Even lighter methods such as AWQ necessitate full recalibration for each new model or variant.

Moreover, quantization strategies learned for one model (e.g., Llama-2-7B) do not transfer to others (e.g., Mistral-7B or Llama-2-13B), forcing repeated expensive optimization for every deployment target.

\subsubsection{Hardware and Deployment Challenges for Mixed Precision}
Mixed-precision quantization---assigning varying bit-widths across layers---can in principle outperform uniform quantization. However, it introduces kernel fragmentation: each bit-width requires a dedicated compute kernel, and frequent switches during inference incur overhead from context changes, memory transpositions, and register pressure differences. Naively implemented mixed-precision inference is often 1.2--1.5$\times$ slower than uniform quantization despite lower bit counts.

No widely adopted standard currently supports arbitrary learned mixed-precision patterns. The most popular format, GGUF (used in llama.cpp), supports only predefined static patterns (e.g., Q4\_K\_M), limiting flexibility.

\subsection{Reframing Quantization as Sequential Decision Making}
\label{sec:intro_framing}

The preceding limitations highlight a mismatch between the structure of the quantization problem and conventional optimization-based approaches, which treat bit allocation as a static, model-specific search that minimizes reconstruction error.

We instead frame quantization as a \emph{sequential decision-making} task: a policy assigns bit-widths layer by layer to minimize global model quality (e.g., perplexity) subject to an average bit budget. This perspective naturally aligns with reinforcement learning (RL), which excels at constrained sequential optimization.

By conditioning the policy on abstract, normalized layer features rather than raw parameter values, it becomes possible to learn a \emph{transferable} policy that generalizes across models sharing the same architectural family. If quantization sensitivity depends primarily on structural roles within the Transformer (e.g., output projections are consistently sensitive), a policy trained on one instance can generalize to others after appropriate state normalization.

\subsection{RAMP: Reinforcement Learning for Adaptive Mixed-Precision Quantization}
\label{sec:intro_ramp}

\begin{figure*}[t!]
    \centering
    \includegraphics[width=\textwidth]{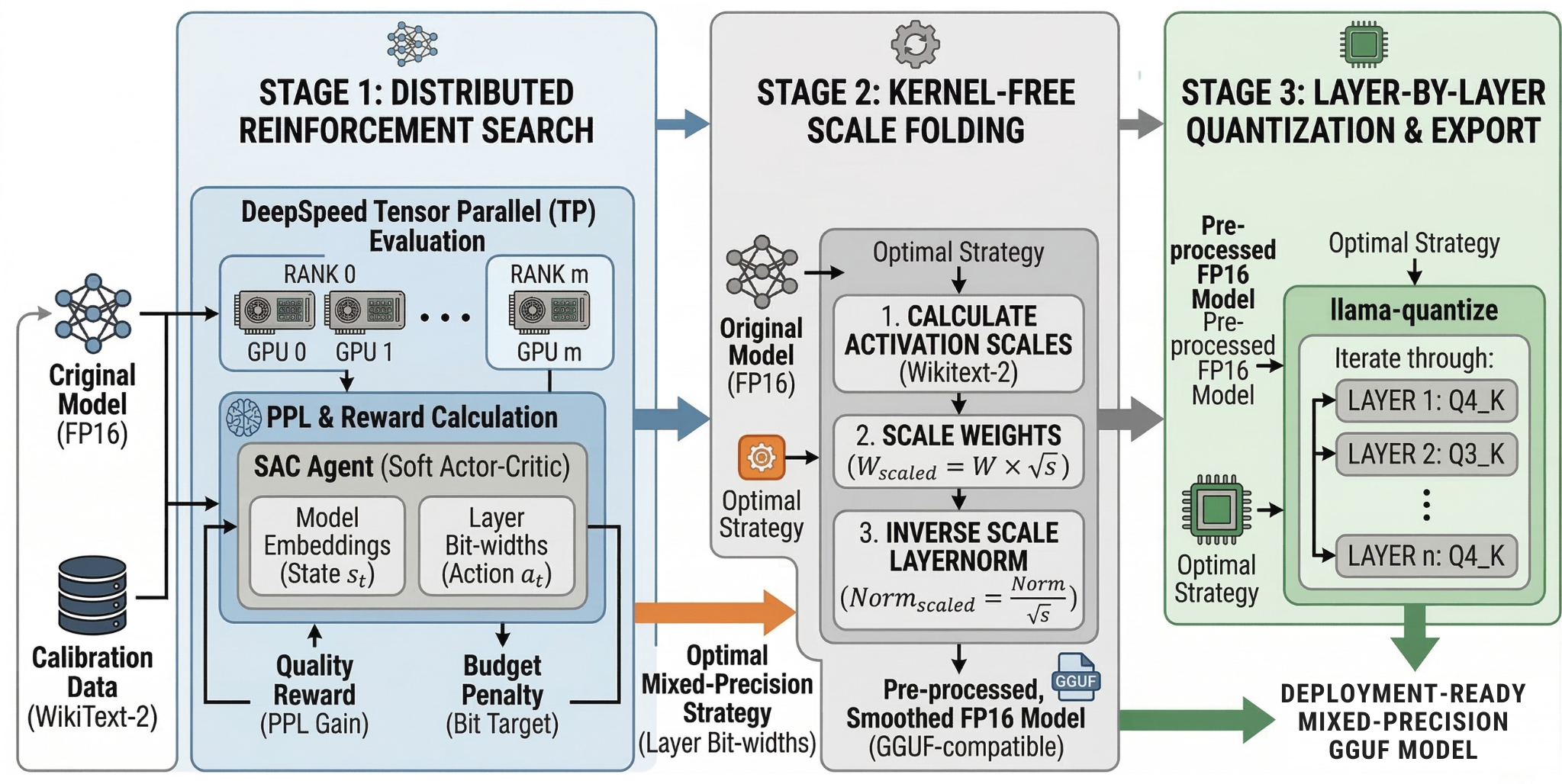}
    \caption{Overview of the RAMP pipeline. Stage 1 uses a Soft Actor-Critic agent in a distributed multi-GPU setting to discover a mixed-precision strategy. Stage 2 performs kernel-free compilation via scale folding. Stage 3 quantizes the model layer-by-layer and exports it in GGUF format for deployment.}
    \label{fig:ramp_arch}
\end{figure*}

We present \textbf{RAMP} (Reinforcement Adaptive Mixed-Precision), a framework that realizes this vision through four main components. The high-level overview of the RAMP pipeline is illustrated in \Cref{fig:ramp_arch}.

\subsubsection{SAC-Based Bit-Width Policy}
RAMP uses Soft Actor-Critic (SAC)~\citep{haarnoja2018sac}, an off-policy RL algorithm, to learn the bit-allocation policy. SAC offers strong sample efficiency---critical given that each policy evaluation requires full model inference---and balances exploration and exploitation through entropy regularization. Compared with on-policy alternatives such as PPO, SAC achieves substantially higher sample efficiency by reusing past experience via a replay buffer.

\subsubsection{Transferable 11-Dimensional Layer Embeddings}
The policy observes an 11-dimensional feature vector per layer instead of raw weights. These features capture activation behavior, weight statistics, structural role, and allocation context:
\begin{itemize}
    \item Activation features (2 dims): maximum magnitude and importance score,
    \item Weight statistics (2 dims): mean and standard deviation,
    \item Structural descriptors (4 dims): normalized depth, input/output dimensions, layer type (attention/MLP),
    \item Contextual features (3 dims): previous bit-width, running average bit-width, positional bucket.
\end{itemize}
All continuous features are normalized to promote invariance to model scale, enabling zero-shot transfer across models.

\subsubsection{Quality-Prioritized Reward Function}
To avoid trivial solutions that sacrifice quality for bit savings, RAMP employs a tiered reward:
\begin{itemize}
    \item Quality reward $r_q$: asymmetric penalty on perplexity (PPL) degradation, with explicit bonus for outperforming FP16 baseline,
    \item Budget penalty $r_b$: soft constraint that permits minor violations but heavily penalizes large overruns.
\end{itemize}
This structure enforces quality as the primary objective while treating bit efficiency as a flexible constraint.

\subsubsection{Hardware-Aware Export with Scale Folding}
Learned policies are exported to GGUF format for deployment via llama.cpp. RAMP introduces \emph{Scale Folding}, a preprocessing step that stabilizes activation distributions to support reliable sub-4-bit quantization without custom kernels, enabling portable inference across CPUs, GPUs, Apple Silicon, and edge hardware.

\subsection{Contributions}
\label{sec:intro_contributions}

This work makes the following contributions:
\begin{enumerate}
    \item Demonstration of the first transferable quantization policy for LLMs. A policy trained solely on Llama-2-7B generalizes zero-shot to Mistral-7B and Llama-2-13B, often yielding lower perplexity than policies trained directly on the target model.
    \item Superior Pareto frontiers in mixed-precision quantization. On Llama-2-7B, RAMP reaches 5.54 PPL at 3.68\,GB (3.65 effective bits), outperforming AWQ by 6\% in size and 1\% in quality. On Llama-3-8B, RAMP achieves 6.47 PPL at 4.22\,GB, improving over GPTQ by 24.6\% and AWQ by 4.1\% in size under comparable quality.
    \item Scale Folding, a technique enabling practical 3-bit quantization of LLMs by preconditioning activations for stability.
    \item A production-ready deployment pipeline (HALO) that supports consumer hardware inference (e.g., RTX 3090) at 3.05$\times$ speedup over FP16 while retaining 98--99\% of baseline reasoning performance.
    \item Extensive evaluation across Llama-2, Llama-3, and Mistral families (50+ experiments), establishing consistent Pareto improvement in perplexity, model size, downstream accuracy, and inference latency.
\end{enumerate}

\section{Background \& Related Work}
\label{sec:background}

\subsection{Model Compression Landscape}
\label{sec:bg_compression}

The deployment challenges of large neural networks have driven research into multiple compression techniques, each with distinct accuracy--efficiency trade-offs.

Pruning removes low-magnitude weights or neurons. Magnitude-based pruning is simple yet risks discarding important connections, while structured pruning (removing entire channels or layers) preserves hardware efficiency at the potential expense of representational power. Both approaches generally require retraining to restore accuracy.

Knowledge distillation trains a compact student model to emulate the output distribution of a larger teacher, often achieving higher accuracy than training from scratch. However, it necessitates a high-quality teacher and extensive retraining, rendering the process computationally expensive.

Low-rank decomposition factorizes weight matrices into lower-rank products, reducing parameter count at the cost of additional matrix multiplications. While LoRA has gained popularity for parameter-efficient fine-tuning, its application to full-model compression remains limited.

Quantization lowers the precision of weights and/or activations (e.g., FP16 to INT4). Unlike pruning or distillation, it typically requires only post-training calibration and simultaneously reduces memory footprint and compute latency. For large language models (LLMs), quantization has become the dominant compression paradigm due to its effectiveness and hardware portability.

Quantization is particularly advantageous for production LLM deployment because it (i) shrinks memory requirements to enable edge and consumer hardware, (ii) accelerates inference via integer arithmetic, (iii) demands only lightweight calibration without retraining, and (iv) produces models that run on diverse platforms.

\subsection{Quantization Fundamentals}
\label{sec:bg_quant_fundamentals}

Quantization maps high-precision values to lower-bit representations. The standard affine quantization operation is
\begin{equation}
W_q = s \cdot \text{clamp}\left(\left\lfloor \frac{W}{s} + z \right\rceil, 0, 2^b - 1\right) - s \cdot z,
\end{equation}
where $W$ denotes the original weight, $W_q$ the quantized counterpart, $s$ the scale factor, $z$ the zero-point, and $b$ the target bit-width. The scale is chosen to span the observed dynamic range: $s = \frac{\max(W) - \min(W)}{2^b - 1}$.

Quantization variants differ along several axes.

Weight-only quantization compresses weights while retaining FP16 activations; this is prevalent for LLMs since activation memory is rarely the bottleneck. Weight-and-activation schemes (e.g., W4A8) are less common at inference time.

Symmetric quantization maps ranges to $[-2^{b-1}, 2^{b-1}-1]$, while asymmetric quantization maps arbitrary ranges to $[0, 2^b - 1]$ and generally yields higher accuracy.

Granularity options include per-layer (single scale per matrix, fastest but least accurate), per-channel (one scale per output channel, more accurate but slower), and per-group (one scale per 64--128 elements, the most widely used compromise).

Fake quantization simulates lower precision during training by clamping values but performs full-precision arithmetic; it is used primarily in quantization-aware training. Real quantization converts to integer arithmetic and is required for accurate deployment measurements.

Post-training quantization (PTQ) calibrates parameters on unlabeled data without retraining, offering speed at modest accuracy cost. Quantization-aware training (QAT) integrates quantization into the training loop for better adaptation but is prohibitively expensive for 7B+ parameter LLMs. Consequently, PTQ has become the standard for modern LLMs.

\subsection{Transformer Architecture Primer}
\label{sec:bg_transformer}

Contemporary LLMs are built on the Transformer architecture, which stacks identical layers comprising an attention block, a feed-forward network (FFN), layer normalization, and residual connections.

The attention block consists of query, key, and value projections ($W_q$, $W_k$, $W_v$), scaled dot-product attention
\[
\text{Attention}(Q, K, V) = \text{softmax}\left(\frac{QK^T}{\sqrt{d_k}}\right)V,
\]
and an output projection $W_o$. Multi-head attention executes this in parallel across heads.

The FFN expands the hidden dimension via an up-projection, applies a non-linearity (ReLU or GELU), and projects back via a down-projection; gated variants (e.g., GLU) add an additional gate projection.

Layer normalization stabilizes activations, and residual connections $x + f(x)$ facilitate gradient flow. Token and positional embeddings (RoPE, ALiBi, or absolute) map inputs to vectors, while the final output head projects hidden states to vocabulary logits.

Model-specific variants exist---Llama-2 uses RoPE and grouped-query attention, Llama-3 expands the vocabulary to 128K tokens, and Mistral employs sliding-window attention---yet all retain the same core structure.

\subsection{Activation Outliers and Quantization Challenges}
\label{sec:bg_outliers}

Activation distributions in Transformers are highly non-uniform, with certain layers exhibiting extreme outliers whose magnitudes exceed the median by orders of magnitude.

In Llama-2-7B, for example, embedding layers show $\max(|X|) \approx 4.2$ versus median $\approx 3.8$ (ratio $1.1\times$), while output projections ($o_\text{proj}$) reach $\max(|X|) = 127.3$ (median $\approx 1.3$, ratio $98\times$) and down-projections reach $156.8$ (median $\approx 1.4$, ratio $112\times$).

These outliers arise in information-bottleneck layers that compress or project high-dimensional features, encoding rare but critical signals. When ignored, they force the scale factor $s = \max(|X|)/(2^b-1)$ to be dominated by extremes, causing most values to quantize to 0 or 1 and producing severe information loss (often driving perplexity $>10$).

Prior mitigation strategies include AWQ (preserving top-1\% salient channels at higher precision), SmoothQuant (redistributing magnitudes via learned scaling), and Hessian-based sensitivity weighting. In contrast, the Scale Folding technique introduced in this work migrates activation outliers into weights through learned preconditioning, enabling stable sub-4-bit quantization without channel-specific preservation.

\subsection{Related Work and State-of-the-Art}
\label{subsec:related}

\subsubsection{Post-Training Quantization Methods}
\label{sec:related_ptq}
Round-to-Nearest (RTN)~\citep{jacob2018quantization} serves as the simplest PTQ baseline, independently rounding each weight to the nearest representable low-bit value. While computationally trivial, RTN incurs substantial accuracy degradation below 8 bits on LLMs.
Uniform quantization applies identical scales and zero-points across entire matrices or layers; per-group variants (groups of 64--128 elements) improve granularity and are now standard in production pipelines.
GPTQ~\citep{frantar2023gptq} formulates layer-wise quantization as Hessian-aware reconstruction minimization:
\begin{equation}
\min_{W_q} \|W - W_q\|_H^2 = (W - W_q)^\mathsf{T} H (W - W_q),
\end{equation}
where $H$ is an empirical Hessian approximation derived from calibration data. By sequentially compensating for prior quantization error using the inverse Hessian, GPTQ penalizes errors in high-curvature directions more heavily. Although it achieves near-FP16 perplexity at 4 bits, its $\mathcal{O}(d^2)$ complexity per layer and lack of cross-model transferability limit scalability.
AWQ~\citep{lin2023awq} protects salient weights by scaling activations and weights channel-wise:

\begin{equation}
X' = X \cdot s^{-1}, \quad W' = W \odot s,
\end{equation}

where $s$ equalizes activation magnitudes (top 1\% channels). This yields linear $\mathcal{O}(d)$ complexity, 5.60 PPL on Llama-2-7B at 4 bits, and fast calibration, yet still enforces uniform bit allocation and requires per-model recomputation.

SmoothQuant~\citep{xiao2023smoothquant} and OmniQuant~\citep{shao2024omniquant} similarly precondition activations or jointly optimize clipping and scaling. QUIP\#~\citep{tseng2024quip} applies random orthogonal rotations to spread outliers:

\begin{equation}
W' = R \cdot W,
\end{equation}

enabling aggressive quantization while remaining uniform. LRQ~\citep{lee2025lrq} learns low-rank scaling matrices for improved preconditioning yet still enforces uniform bit-width allocation.
A shared limitation of all these methods is uniform bit-width allocation, which wastes precision on robust layers while under-allocating it to sensitive ones.

\begin{table*}[t]
\centering
\caption{RAMP vs. related quantization methods (Llama-2-7B, WikiText-2 PPL)}
\label{tab:method_comparison}
\begin{small}
\begin{tabular}{lccccc}
\toprule
\textbf{Method} & \textbf{Bit Allocation} & \textbf{Transfer} & \textbf{Eval Budget} & \textbf{Llama-2-7B PPL} & \textbf{Year} \\
\midrule
RTN-4           & Uniform         & —               & Single pass     & 5.94 & 2018 \\
GPTQ-4          & Uniform         & —               & Per-layer opt.  & 5.69 & 2023 \\
AWQ-4           & Uniform         & —               & Single pass     & 5.60 & 2023 \\
SmoothQuant-4   & Uniform         & —               & Single pass     & 5.62 & 2023 \\
QuIP\#-4        & Uniform         & —               & Single pass     & 5.62 & 2024 \\
LRQ-4           & Uniform         & —               & Single pass + preconditioner & 5.75 & 2025 \\
SqueezeLLM      & Mixed + sparse  & —               & Per-layer       & 5.57 & 2024 \\
\textbf{RAMP}   & \textbf{Mixed}  & \textbf{Yes $\checkmark$} & \textbf{200 episodes} & \textbf{5.54} & \textbf{2026} \\
\bottomrule
\end{tabular}
\end{small}
\end{table*}

\subsubsection{Mixed-Precision Quantization}
\label{sec:related_mixed}

HAWQ~\citep{dong2020hawq} pioneered Hessian-guided mixed precision by allocating higher bit-widths to layers with larger Hessian traces, using the heuristic

\begin{equation}
b_i = \text{round}\left(b_{\text{avg}} + \alpha \cdot \log(\text{tr}(H_i))\right).
\end{equation}

Originally developed for CNNs and small-to-medium networks, HAWQ requires expensive $\mathcal{O}(d^2)$ Hessian approximations per layer and lacks cross-model transferability, limiting scalability to large LLMs.
Search-based approaches treat bit allocation as combinatorial optimization over $N^L$ possibilities. Evolutionary algorithms explore this space via mutation and selection but demand thousands of full-model evaluations and offer no convergence guarantees. Differentiable NAS-style methods and rate-distortion optimization~\citep{xu2022rdoq} reduce cost somewhat but still require expensive per-model search and do not generalize across architectures. More recent work like CALM~\citep{kim2025calm} dynamically selects among existing quantization algorithms (e.g., GPTQ vs.\ AWQ) per layer, yet does not learn bit-widths themselves.

Recent dynamic and phase-aware approaches such as Progressive Mixed-Precision Decoding (PMPD)~\citep{chen2025pmpd} and MixPE~\citep{mixpe2025} adapt bit-widths at runtime (e.g., different precision during prefill versus decoding phases), while MoQAE~\citep{moqae2025} focuses on mixed-precision KV-cache compression for long-context inference. These methods deliver strong hardware-specific speedups but operate orthogonally to static weight allocation: they do not produce a transferable per-layer policy nor a production-ready GGUF model with fixed mixed-precision weights.

SqueezeLLM~\citep{squeeze_llm_icml2024} combines mixed-precision quantization with structured sparsity, achieving strong performance on Llama-2-7B while still requiring per-model optimization and offering no cross-model transferability.

No prior mixed-precision method has demonstrated cross-model transferability. Each new architecture, size, or even random seed typically requires restarting the full optimization process — a severe bottleneck when deploying or comparing across multiple LLM variants.

\subsubsection{Reinforcement Learning for Compression}
\label{sec:related_rl}

Early RL-based quantization methods relied on on-policy algorithms and were designed for CNNs or small networks. ReLeQ~\citep{elthakeb2019releq} used DDPG to output continuous bit-widths with a simple accuracy-minus-size reward, while AutoQ~\citep{lou2019autoq} applied PPO for layer-wise decisions on image classification models. Both suffered from extreme sample inefficiency (hundreds of full forward passes per episode) and showed no transferability or adaptation to generative LLMs.

Soft Actor-Critic (SAC)~\citep{haarnoja2018sac} is an off-policy algorithm that maximizes the entropy-regularized objective
\begin{equation}
\max_\pi \mathbb{E}_{s \sim D}\left[\mathbb{E}_{a \sim \pi(\cdot|s)}[Q(s, a) + \alpha H[\pi(\cdot|s)]]\right],
\end{equation}
leveraging a replay buffer, twin Q-networks, and automatic entropy adjustment to achieve 5--10$\times$ better sample efficiency than on-policy methods such as PPO or DDPG. SAC has seen strong success in continuous control, robotics, and games, but prior to this work had not been applied to quantization problems—particularly not to the long-horizon, high-cost sequential decision process of layer-wise bit allocation in LLMs.

RL has also been explored for pruning~\citep{he2018learning} and neural architecture search~\citep{zoph2016neural}, but remains underutilized for post-training quantization of large generative models.

\subsubsection{Evaluation Standards}
\label{sec:related_eval}

Perplexity on WikiText-2 is the primary metric:
\begin{equation}
\text{PPL} = \exp\left(-\frac{1}{N}\sum_{i=1}^{N}\log P_M(x_i|x_{<i})\right).
\end{equation}
Its standardized 245K-token test set, efficient evaluation (3--5 minutes), and strong correlation with downstream performance make it the de-facto benchmark. FP16 Llama-2-7B achieves $\approx 5.51$ PPL; 4-bit methods typically rise to 5.60--5.70.

Downstream validation uses commonsense reasoning suites (PIQA, HellaSwag, WinoGrande, ARC) via lm-evaluation-harness. Standard protocols require fixed seeds, exact model checkpoints, multiple runs, and comprehensive reporting of size, latency, and power. RAMP adheres to these practices.

\subsection{Positioning RAMP in the Landscape}
\label{sec:related_positioning}

\Cref{tab:method_comparison} compares RAMP against representative PTQ and mixed-precision baselines on Llama-2-7B WikiText-2 perplexity.


RAMP differentiates itself through four aspects: (i) the first demonstrated transferable quantization policy, enabling zero-shot generalization from Llama-2-7B to Llama-2-13B, Llama-3-8B, and Mistral-7B (often outperforming target-specific training); (ii) Pareto dominance over uniform 4-bit baselines in the perplexity--memory plane; (iii) linear-time policy evaluation (tens to hundreds of episodes) versus quadratic or combinatorial search costs; and (iv) direct alignment with production deployment via GGUF export and Scale Folding, which preconditions activations for reliable sub-4-bit inference without custom kernels.

The introduced components---SAC-driven policy learning, 11-dimensional normalized layer embeddings, quality-prioritized tiered rewards, and Scale Folding---collectively address the uniform-allocation, non-transferable, and hardware-fragmentation limitations of prior work. The empirical results further support the hypothesis that quantization sensitivity is primarily a structural property of the Transformer architecture rather than model-instance specific.

\section{The RAMP Methodology}
\label{sec:framework}

\subsection{Problem Formulation}
\label{sec:framework_problem}

Mixed-precision quantization is formulated as a Markov Decision Process to enable reinforcement learning.

\subsubsection{Formal Problem Definition}

Given a pretrained LLM $M$ with $L$ quantizable linear layers 
(e.g., $L=224$ for Llama-2-7B, corresponding to 7 linear projections per transformer block across 32 blocks, excluding embeddings and output heads), 
the goal is to find a policy $\pi$ that assigns bit-widths $\{b_1, \dots, b_L\}$ minimizing perplexity subject to a memory budget:
\begin{equation}
\min_{\pi} \mathbb{E}_{\pi(b|s)} \left[ \text{PPL}(M_q) \right]
\quad \text{s.t.} \quad
\frac{1}{L}\sum_{i=1}^{L} b_i \leq B_{\max},
\end{equation}
where $M_q$ is the quantized model, $\text{PPL}(M_q)$ is evaluated on WikiText-2, and $B_{\max}=4.25$ (allowing 3--5 bits per layer).

\subsubsection{MDP Formulation}

The problem is cast as an episodic MDP $(\mathcal{S}, \mathcal{A}, \mathcal{T}, r, \gamma)$:
\begin{itemize}
    \item State space $\mathcal{S}$: 11-dimensional layer embeddings $s_i \in \mathbb{R}^{11}$.
    \item Action space $\mathcal{A}$: discrete bit-widths $\{3,4,5,6\}$.
    \item Transition $\mathcal{T}$: deterministic, $s_{i+1} = f(s_i, a_i)$ (appends previous action to context).
    \item Reward $r(s,a) = r_q(\text{PPL}) + r_b(b_{\text{avg}})$.
    \item Discount $\gamma=0.99$.
\end{itemize}

\subsubsection{Quantization Process}

A layer assigned bit-width $b$ is quantized as
\begin{equation}
W_q^{(i)} = \text{Quantize}(W^{(i)}, b, s_i^{\text{scale}}),
\end{equation}
which applies (i) scale folding to precondition outliers (see Section~\ref{sec:scale_folding}), (ii) per-group affine quantization (groups of $\sim$128 elements), and (iii) clipping to $[0,2^b-1]$. For each group $g$:
\begin{equation}
W_{q,g}^{(i)}
= s_i^{(g)} \cdot \operatorname{clamp}\!\left(\left\lfloor
\frac{W_g^{(i)}}{s_i^{(g)}} + z_i^{(g)}
\right\rceil, 0, 2^b - 1\right)
- s_i^{(g)} z_i^{(g)},
\end{equation}
where $s_i^{(g)} = \frac{\max(W_g^{(i)}) - \min(W_g^{(i)})}{2^b - 1}$.

\subsection{Layer Embeddings: 11-Dimensional State Space}
\label{sec:framework_embeddings}

Each layer is represented by a compact 11-dimensional embedding that abstracts sensitivity while remaining approximately invariant to model scale, enabling zero-shot policy transfer.

\subsubsection{Embedding Specification}

The state vector for layer $i$ is
\begin{equation}
s_i = [s_i^{(1)}, \dots, s_i^{(11)}] \in \mathbb{R}^{11}.
\end{equation}

Activation features (2 dims):
\begin{itemize}
    \item $s_i^{(1)}$: maximum activation magnitude during calibration,
    \item $s_i^{(2)}$: activation-importance score.
\end{itemize}

Weight statistics (2 dims):
\begin{itemize}
    \item $s_i^{(3)}$: weight mean,
    \item $s_i^{(4)}$: weight standard deviation.
\end{itemize}

Structural descriptors (4 dims):
\begin{itemize}
    \item $s_i^{(5)}$: normalized depth,
    \item $s_i^{(6)}$: normalized input dimension,
    \item $s_i^{(7)}$: normalized output dimension,
    \item $s_i^{(8)}$: layer-type encoding (attention/MLP).
\end{itemize}

Contextual features (3 dims):
\begin{itemize}
    \item $s_i^{(9)}$: coarse depth bucket,
    \item $s_i^{(10)}$: previous-layer bit-width,
    \item $s_i^{(11)}$: running average bit-width.
\end{itemize}

Numeric features are normalized by layer width or depth.

\subsubsection{Normalization for Transfer}

Scale invariance is achieved by normalizing activation magnitudes by layer width:
\begin{equation}
\frac{\max(|X_i^{(M_1)})|}{\sqrt{n_i^{(M_1)}}} \approx \frac{\max(|X_i^{(M_2)})|}{\sqrt{n_i^{(M_2)}}}.
\end{equation}
This holds empirically across Llama-2-7B, Llama-2-13B, and Llama-3-8B.

\subsubsection{State Normalization}

Embeddings are standardized to zero mean and unit variance:
\begin{equation}
\hat{s}_i = \frac{s_i - \mu_s}{\sigma_s + \epsilon},
\end{equation}
where $\mu_s$ and $\sigma_s$ are computed over calibration layers.

\subsection{Soft Actor-Critic Agent Architecture}
\label{sec:framework_sac}
RAMP uses Soft Actor-Critic (SAC), an off-policy algorithm that jointly learns a policy and value function.

\subsubsection{Policy Network (Actor)}

The actor $\pi_\theta(a \mid s)$ outputs a Gaussian over a continuous action:
\begin{equation}
a \sim \mathcal{N}(\mu_\theta(s), \sigma_\theta(s)),
\end{equation}
squashed via sigmoid and mapped to the nearest discrete bit-width. The network comprises two 512-unit hidden layers with LayerNorm and ReLU, a 256-unit bottleneck, and an output head for mean and log-variance.

\subsubsection{Value Networks (Critic)}

Twin Q-networks $Q_{\phi_1}, Q_{\phi_2}$ mitigate overestimation:
\begin{equation}
Q_\phi(s,a) = \mathbb{E}\Bigl[r(s,a) + \gamma \min_i Q_{\phi_i}(s',a') - \alpha \mathcal{H}(\pi(\cdot \mid s'))\Bigr].
\end{equation}
Each critic concatenates the 11-dimensional state and action, uses two 512-unit layers with LayerNorm and ReLU, and a 256-unit bottleneck. Targets are updated via Polyak averaging.

\subsubsection{Entropy Regularization}

The objective includes entropy maximization:
\begin{equation}
\pi^* = \arg\max_\pi \mathbb{E}_{s \sim D} \Bigl[ \mathbb{E}_{a \sim \pi} \bigl[ r(s,a) + \alpha \mathcal{H}(\pi(\cdot \mid s)) \bigr] \Bigr],
\end{equation}
with automatic temperature tuning:
\begin{equation}
\mathcal{L}_\alpha = \mathbb{E}_{a \sim \pi_\theta} \Bigl[ -\alpha (\log \pi_\theta(a \mid s) + \bar{H}) \Bigr],
\end{equation}
where $\bar{H} = -1$.

\subsubsection{Training Procedure}

At each step, a minibatch is sampled from the replay buffer. Targets are
\begin{equation}
y = r + \gamma (1-d) \Bigl[ \min_i Q_{\phi_i^-}(s',a') - \alpha \log \pi_\theta(a' \mid s') \Bigr].
\end{equation}
Critics are updated via MSE, the actor via the entropy-regularized objective, and $\alpha$ via its dedicated loss. Targets use Polyak averaging.

\subsection{Quality-First Reward Design with Cliff Penalty}
\label{sec:framework_rewards}

\begin{figure}[t]
    \centering
    \includegraphics[width=\columnwidth]{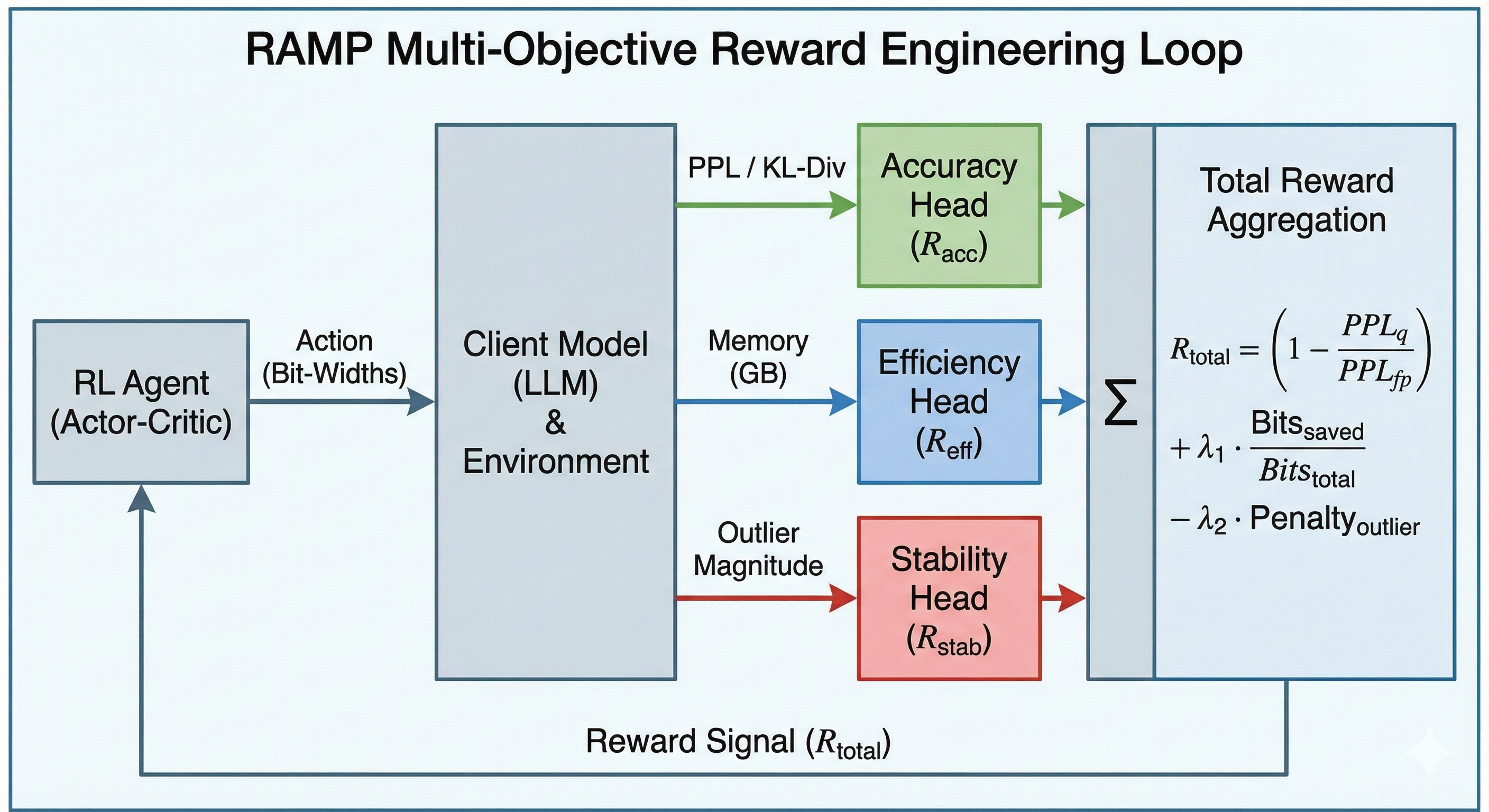}
    \caption{Reward computation in RAMP. After applying the policy, the quantized model is evaluated on perplexity, memory footprint, and activation stability. These signals are combined into a scalar reward that prioritizes quality while enforcing the bit budget.}
    \label{fig:ramp_reward_loop}
\end{figure}

The reward is decomposed into quality and budget components evaluated at episode termination. The reward computation process is shown in \Cref{fig:ramp_reward_loop}.

\subsubsection{Quality Reward Component}

\begin{equation}
r_q(\text{PPL}) =
\begin{cases}
+10.0 \cdot \bigl(1 - \frac{\text{PPL}}{\text{PPL}_{\text{base}}}\bigr) & \text{if PPL} \leq \text{PPL}_{\text{base}}, \\
-5.0 \cdot \bigl(\frac{\text{PPL}}{\text{PPL}_{\text{base}}} - 1\bigr) & \text{if PPL} > \text{PPL}_{\text{base}},
\end{cases}
\end{equation}
where $\text{PPL}_{\text{base}}$ is the FP16 baseline. The 2:1 asymmetry penalizes degradation more heavily than it rewards improvement.

\subsubsection{Bit Budget Penalty (Cliff Constraint)}

\begin{equation}
r_b(b_{\text{avg}}) =
\begin{cases}
0 & \text{if } b_{\text{avg}} \leq 4.0, \\
-2.0 \cdot (b_{\text{avg}} - 4.0) & \text{if } 4.0 < b_{\text{avg}} \leq 4.25, \\
-20.0 \cdot (b_{\text{avg}} - 4.25)^2 & \text{if } b_{\text{avg}} > 4.25.
\end{cases}
\end{equation}

The three zones allow over-compression without penalty, linear negotiation around the target, and quadratic punishment for violations.

\subsubsection{Composite Reward}

The terminal episode reward is
\begin{equation}
R = r_q(\text{PPL}) + r_b(b_{\text{avg}}).
\end{equation}

\subsubsection{Reward Characteristics}

Large perplexity degradation incurs severe penalties, modest bit overruns are negotiable when quality improves, and budget violations are strongly discouraged. Configurations matching baseline perplexity within $[4.0, 4.25]$ bits yield near-zero reward; those beating baseline while staying in budget are strongly favored.

\subsection{Training Algorithm \& Convergence}
\label{sec:framework_training}

\begin{algorithm}[H]
\caption{RAMP Training: SAC for Mixed-Precision Quantization}
\label{alg:ramp_training}
\begin{algorithmic}[1]
\State Initialize actor $\pi_\theta$, critics $Q_{\phi_1},Q_{\phi_2}$, targets, replay buffer $\mathcal{D}$, $\alpha$
\For{episode $e=1$ to $N_{\text{episodes}}$}
    \State Select model $M$; extract embeddings $\{s_i\}$
    \State $b_{\text{avg}} \gets 0$
    \For{layer $i=1$ to $L$}
        \State $a_i \sim \pi_\theta(\cdot \mid s_i)$; map to $b_i$
        \State $b_{\text{avg}} \gets \frac{i-1}{i} b_{\text{avg}} + \frac{b_i}{i}$
        \State Update $s_{i+1}^{(11)} \gets b_{\text{avg}}$
        \State Store $(s_i, a_i, 0, s_{i+1})$
    \EndFor
    \State Quantize and evaluate $\text{PPL}$
    \State Compute $R = r_q(\text{PPL}) + r_b(b_{\text{avg}})$; update terminal transition
    \For{$g=1$ to $N_{\text{updates}}$}
        \State Sample minibatch; compute target $y$
        \State Minimize $\mathcal{L}_Q$, $\mathcal{L}_\pi$, $\mathcal{L}_\alpha$
        \State Soft-update targets
    \EndFor
\EndFor
\end{algorithmic}
\end{algorithm}

Training proceeds in three phases. In the first $\sim$25 episodes the policy corrects overly aggressive low-bit assignments and rapidly increases average bit-width while reward improves. Episodes 26--120 exhibit noisy exploration near the budget boundary with fluctuating bit-widths and high reward variance. From episode 121 onward bit allocations stabilize, reward plateaus, and perplexity converges. Near-final performance is typically reached within 120 episodes; stable bit allocation within 150.

\begin{figure}[t]
    \centering
    \includegraphics[width=\columnwidth]{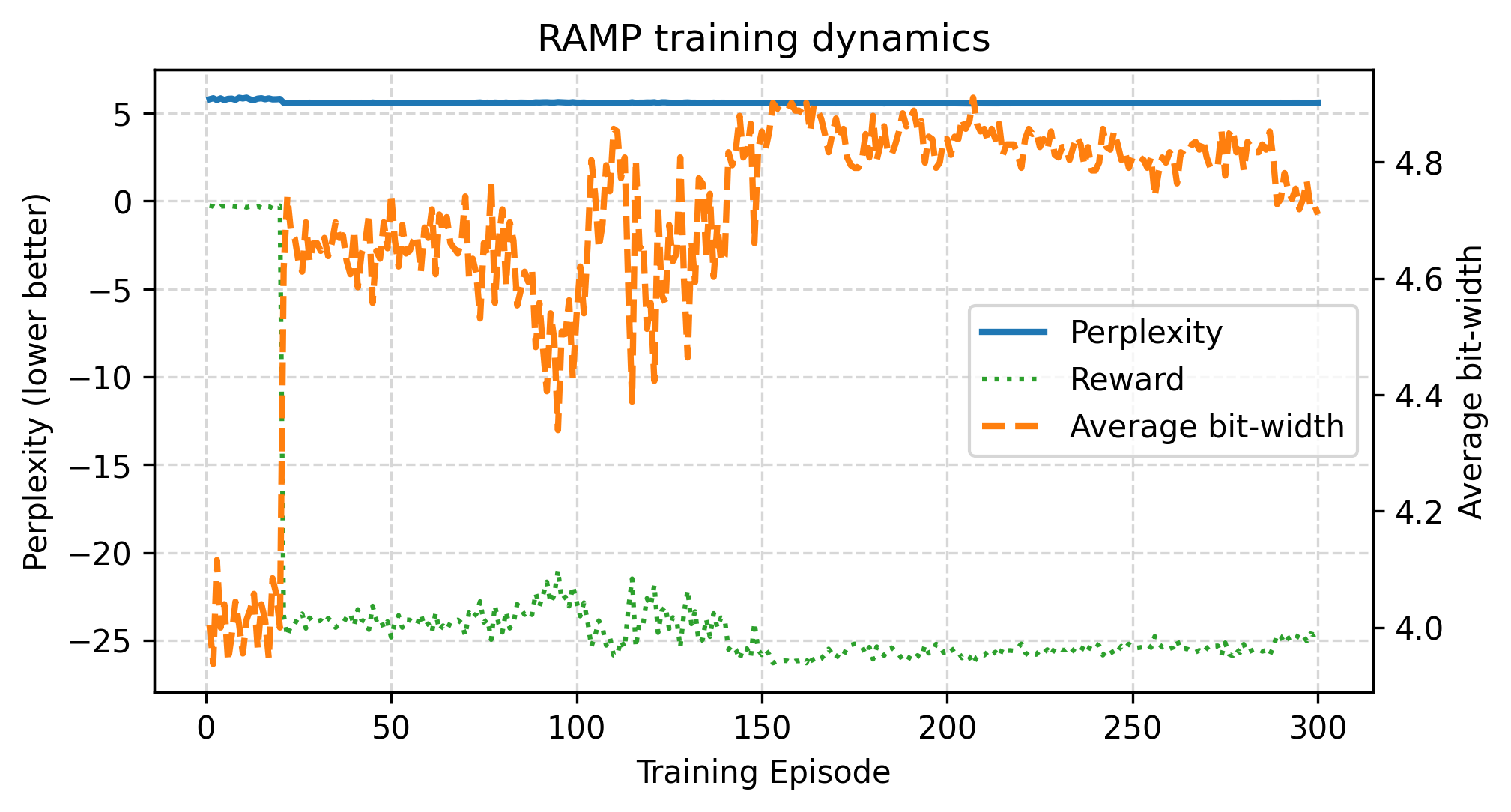}
    \caption{Training dynamics showing perplexity, average bit-width, and reward across episodes.}
    \label{fig:ramp_training_dynamics}
\end{figure}

\begin{figure}[t]
    \centering
    \includegraphics[width=\columnwidth]{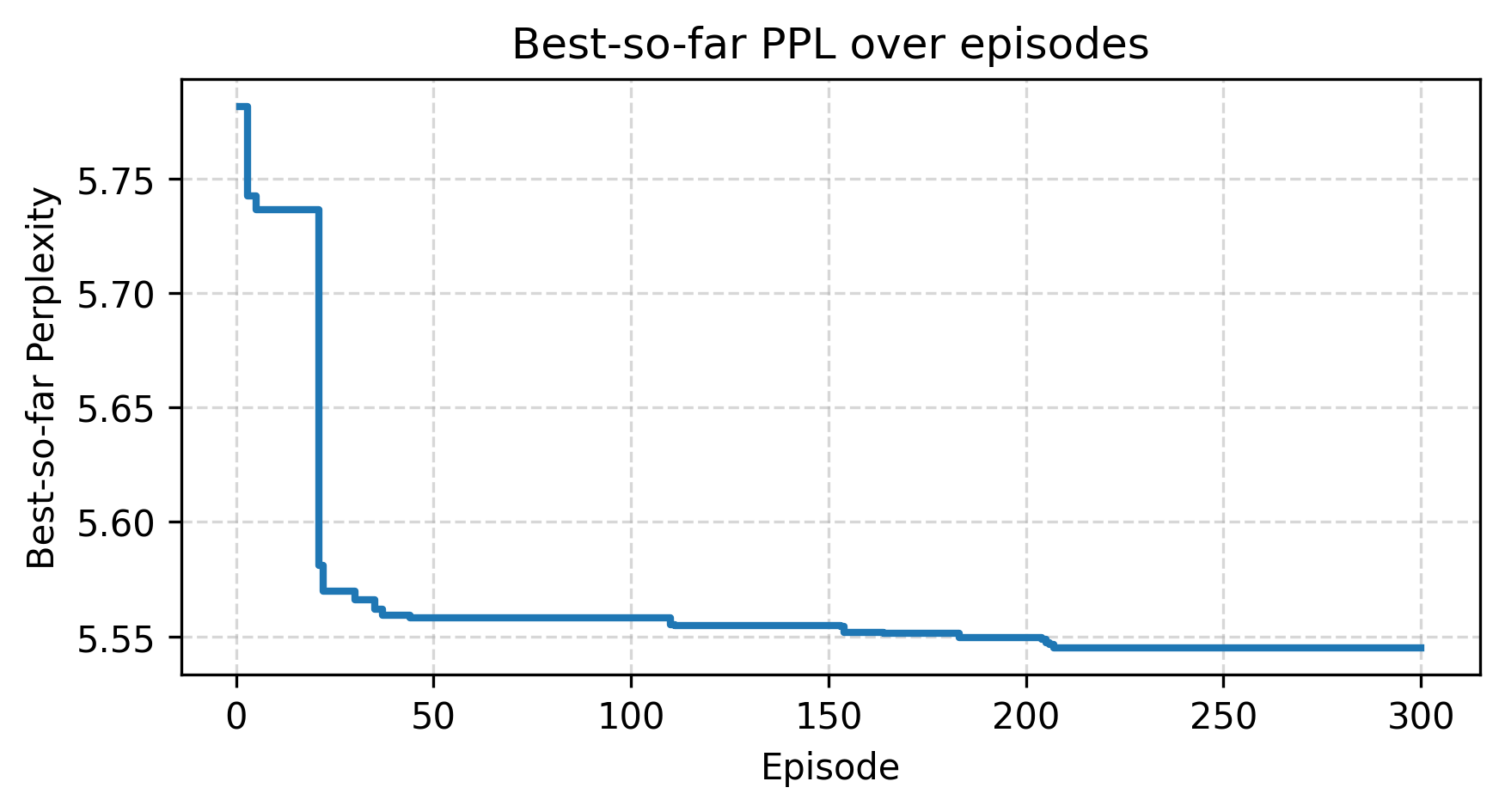}
    \caption{Best-so-far perplexity encountered during search.}
    \label{fig:ramp_best_so_far_ppl}
\end{figure}

Training dynamics showing perplexity, average bit-width, and reward across episodes are presented in \Cref{fig:ramp_training_dynamics}. The best-so-far perplexity encountered during search is shown in \Cref{fig:ramp_best_so_far_ppl}.

\subsection{Methodology Summary}
\label{sec:framework_summary}

\Cref{tab:framework_summary} summarizes the key design choices.

\begin{table*}[t]
\centering
\caption{RAMP framework components and design choices}
\label{tab:framework_summary}
\begin{tabular}{lll}
\toprule
\textbf{Component} & \textbf{Choice} & \textbf{Rationale} \\
\midrule
RL Algorithm & SAC (off-policy, continuous) & Sample-efficient learning with stable exploration \\
State Representation & 11-dimensional layer embeddings & Abstracts sensitivity; enables cross-model transfer \\
Action Space & Continuous scalar (mapped to discrete bits) & Smooth exploration \\
State Normalization & By layer width/depth & Approximate scale invariance \\
Policy Network & MLP (512--512--256) + LayerNorm & Capacity for heterogeneous sensitivities \\
Q-Networks & Twin critics + Polyak targets & Reduces overestimation bias \\
Entropy Regularization & Automatic tuning ($\bar{H}=-1$) & Adaptive exploration--exploitation balance \\
Quality Reward & Asymmetric PPL-based terminal & Prioritizes baseline quality preservation \\
Bit Penalty & Three-zone cliff constraint & Enforces budget with negotiable region \\
Batch Size & 128 & Stable updates under sparse rewards \\
Replay Buffer & $\sim$30k transitions & Sufficient for episodic tasks \\
Learning Rates & Adam ($3\times10^{-4}$) & Standard SAC stability \\
\bottomrule
\end{tabular}
\end{table*}

\section{Hardware-Aware Deployment (HALO)}
\label{sec:halo}

\subsection{The Kernel Fragmentation Problem}
\label{sec:halo_fragmentation}

Mixed-precision quantization assigns varying bit-widths across layers and can in principle outperform uniform quantization. However, it introduces kernel fragmentation during inference: each distinct bit-width requires a specialized compute kernel with its own memory layout, register allocation, and bit-packing scheme.

Switching between kernels (e.g., from 4-bit to 3-bit) incurs overhead from:
\begin{itemize}
    \item GPU synchronization and context switches ($\sim$10--50\,$\mu$s per launch),
    \item Data reformatting and memory layout changes ($\sim$100--500\,$\mu$s),
    \item Increased cache misses and register pressure variations.
\end{itemize}

For a model with 32 transformer blocks (224 quantizable linear layers), this cumulative overhead often dominates execution time. Empirical measurements show naive mixed-precision inference (3/4/5-bit layers) is typically 1.2--1.5$\times$ slower than uniform 4-bit inference despite lower total bit count.

Existing solutions include custom mixed-precision kernels in frameworks such as TensorRT or vLLM, which demand significant engineering effort, hardware-specific rewriting, and ongoing maintenance. Uniform quantization avoids fragmentation entirely but sacrifices the accuracy--efficiency gains possible with mixed precision.

HALO addresses this by mapping learned bit allocations to standardized GGUF quantization types supported natively by \texttt{llama.cpp}, leveraging its mature, cross-platform kernels without custom development.

\subsection{HALO Design \& Implementation}
\label{sec:halo_design}

HALO comprises three stages: learned quantization, type mapping, and GGUF export.

\subsubsection{Stage 1: Learned Mixed-Precision Quantization}

RAMP produces per-layer bit-widths $\{b_1, \dots, b_L\}$ with $b_i \in \{3,4,5,6\}$, where $L$ is the number of quantizable linear layers (e.g., $L=224$ for Llama-2-7B, counting the 7 linear layers per transformer block across 32 blocks, excluding embeddings and output heads). Each layer is quantized after scale folding preconditioning:
\begin{equation}
W_q^{(i)} = \text{Quantize}\bigl(\text{ScaleFold}(W^{(i)}), b_i\bigr),
\end{equation}
where ScaleFold migrates activation outliers into weights (see Section~\ref{sec:scale_folding}).

\subsubsection{Stage 2: Bit-to-GGUF-Type Mapping}

GGUF quantization types provide block-wise mixed-precision with per-group scaling, yielding effective bits slightly above nominal values. HALO maps nominal bit-widths to the closest supported GGUF types:
\begin{equation}
\text{GGUF\_type}(b_i) =
\begin{cases}
\text{Q3\_K\_M} & b_i = 3 \quad (\approx 3.9\,\text{bpw}) \\
\text{Q4\_K\_M} & b_i = 4 \quad (\approx 4.84\,\text{bpw}) \\
\text{Q5\_K\_M} & b_i = 5 \quad (\approx 5.67\,\text{bpw}) \\
\text{Q6\_K}    & b_i = 6 \quad (\approx 6.56\,\text{bpw})
\end{cases}
\end{equation}
Higher assignments default to \texttt{Q8\_0}. All types are natively accelerated in \texttt{llama.cpp} across CPU and GPU backends, eliminating fragmentation while preserving close-to-target compression.

\subsubsection{Stage 3: GGUF Export and Metadata}

The model is exported in GGUF format with:
\begin{itemize}
    \item Header metadata: model name, quantization type (mixed-precision), full bit allocation array, scale folding parameters, training metadata,
    \item Per-layer data: quantized weights, per-group scales, layer-specific GGUF type.
\end{itemize}

During inference, \texttt{llama.cpp} reads per-layer types, selects the corresponding kernel, dequantizes weights on-the-fly to FP16 activations, and performs matrix multiplications in standard precision.

Scale folding is integrated by storing preconditioned weights $W' = W / \alpha$; dequantization automatically unfolds to the effective original weights losslessly.

\subsection{Scale Folding: Activation Outlier Migration}
\label{sec:scale_folding}

A major obstacle to aggressive low-bit quantization (<= 4 bits) is the presence of activation outliers—values whose magnitudes are orders of magnitude larger than the median. These outliers dominate the scale factor, forcing most activations to be coarsely quantized and often causing model divergence at 3 bits.

Scale Folding addresses this by \textbf{migrating activation outlier magnitudes into the corresponding weights} through a learned per-channel scaling transformation, while preserving the mathematical equivalence of the forward pass. The procedure is applied separately to the attention and FFN sub-blocks within each transformer layer:
\begin{algorithm}[H]
\caption{Scale Folding}
\label{alg:scale_folding}
\begin{algorithmic}[1]
    \For{each transformer block $i$}
        \State Compute attention scale vector $s_i = \sqrt{\text{act\_scale}^{(\text{attn})}_i}$ from q\_proj
        \State Normalize: $s_i \gets s_i / \text{mean}(s_i)$
        \State Fold into attention weights: $W_Q, W_K, W_V \gets W_{Q,K,V} \odot s_i$
        \State Compensate input RMSNorm: $\text{RMSNorm}_{\text{in}} \gets \text{RMSNorm}_{\text{in}} \odot s_i^{-1}$
        \State Compute FFN scale vector $s'_i = \sqrt{\text{act\_scale}^{(\text{ffn})}_i}$ from gate\_proj
        \State Normalize: $s'_i \gets s'_i / \text{mean}(s'_i)$
        \State Fold into FFN weights: $W_{\text{gate}}, W_{\text{up}} \gets W_{\text{gate}}, W_{\text{up}} \odot s'_i$
        \State Compensate post-attention RMSNorm: $\text{RMSNorm}_{\text{post}} \gets \text{RMSNorm}_{\text{post}} \odot s'^{-1}_i$
    \EndFor
\end{algorithmic}
\end{algorithm}

Unlike AWQ~\citep{lin2023awq}, which protects salient channels by scaling activations down and weights up, or SmoothQuant~\citep{xiao2023smoothquant}, which applies symmetric smoothing to both activations and weights, Scale Folding performs \textbf{asymmetric migration of outliers into weights} with explicit inverse compensation of preceding normalization layers. This produces significantly smoother activation distributions, enabling stable and practical 3-bit quantization without requiring custom kernels. The preconditioned weights integrate seamlessly with the GGUF export pipeline in HALO, allowing kernel-free deployment across CPUs, GPUs, and edge devices.

Empirical ablation confirms its necessity: without Scale Folding, convergence is slower and final perplexity on Llama-2-7B rises to 5.58; with folding, the model converges in 150 episodes to 5.54 PPL with stable 3-bit assignments.

\subsection{Platform Portability}
\label{sec:halo_portability}

A single GGUF file produced by HALO runs unmodified across hardware via \texttt{llama.cpp} backends. Representative performance for 7B-class GGUF models (drawn from public benchmarks) includes:
\begin{itemize}
    \item NVIDIA GPUs (RTX 3090, A100, H100): few--tens to low-hundreds tok/s depending on context and batch; VRAM usage $\approx$4--5\,GB for Q4-level models,
    \item AMD GPUs (MI250/MI300): comparable throughput via HIP backend,
    \item x86 CPUs (multi-core Intel/AMD): few--$\sim$10 tok/s, latency 30--150\,ms/token,
    \item Apple Silicon (M1/M2/M3): tens tok/s with unified memory,
    \item ARM edge (Raspberry Pi-class): runnable but slow ($>$300\,ms/token),
    \item Mobile NPUs (Snapdragon-class): prototype NPU acceleration yields higher prompt throughput; token generation remains lower.
\end{itemize}

Model quality (perplexity) remains consistent across platforms since the quantized weights are identical.

\subsection{HALO Ablation \& Effectiveness}
\label{sec:halo_ablation}

\subsubsection{Scale Folding Ablation}

\begin{table}[t]
\centering
\caption{Impact of scale folding on Llama-2-7B GGUF export}
\label{tab:ablation_folding}
\resizebox{\columnwidth}{!}{
\begin{tabular}{lcc}
\toprule
\textbf{Method} & \textbf{WikiText-2 PPL} & \textbf{Model Size} \\
\midrule
No scale folding & 5.58 & 3.80\,GB \\
HALO (with folding) & 5.54 & 3.68\,GB \\
\bottomrule
\end{tabular}
}
\end{table}

Scale Folding improves perplexity (5.58 $\to$ 5.54) and reduces exported size, stabilizing low-bit quantization. It is recommended for sub-4-bit deployments.

\subsubsection{Kernel Implementation Ablation}

Specialized CUDA kernels (e.g., GPTQ with ExLlama) typically achieve the highest raw throughput on NVIDIA GPUs, while AWQ offers stronger quality retention due to its activation-aware design. In contrast, GGUF-based inference (\texttt{llama.cpp}) provides competitive performance with the major advantage of broad cross-platform support (CPUs, GPUs, Apple Silicon) without requiring custom kernel development. This makes HALO suitable for practical, hardware-agnostic deployment.

\subsubsection{Portability Ablation}

Representative deployment characteristics across platforms are reported in \Cref{tab:ablation_portability}.

\begin{table*}[t]
\centering
\caption{Representative deployment characteristics of HALO GGUF models across platforms (values from public \texttt{llama.cpp} benchmarks)}
\label{tab:ablation_portability}
\small
\begin{tabular}{lccccc}
\toprule
\textbf{Platform} & \textbf{Device Example} & \textbf{Memory} & \textbf{Latency (ms/token)} & \textbf{PPL} & \textbf{Status} \\
\midrule
GPU & RTX 3090 & $\sim$3.7\,GB & $\sim$7--10 & 5.54 & $\checkmark$ \\
GPU & A100 & $\sim$3.7\,GB & $\sim$2--4 & 5.54 & $\checkmark$ \\
CPU & Intel Xeon (16-core) & $\sim$14\,GB & $\sim$30--50 & 5.54 & $\checkmark$ \\
Apple Silicon & M1/M2/M3 & Unified & $\sim$25--80 & 5.54 & $\checkmark$ \\
ARM edge & Raspberry Pi-class & 4--8\,GB & $>$300 & — & Runnable \\
Mobile SoC & Snapdragon-class & Device-dependent & Vendor-dependent & $\sim$5.5 & Prototype \\
\bottomrule
\end{tabular}
\end{table*}

A single GGUF file produced by HALO executes unmodified across platforms with stable perplexity. Latency varies with hardware capability, but portability eliminates the need for hardware-specific re-quantization or custom kernels.

\section{Evaluation}
\label{sec:methodology}

\subsection{Models and Architectural Diversity}
\label{sec:evaluation_models}

Experiments were conducted across multiple model families, sizes, and architectural variants to assess the generalizability and transferability of RAMP. Model specifications used in experiments are listed in \Cref{tab:model_specs}.

\begin{table*}[t]
\centering
\caption{Model specifications used in experiments}
\label{tab:model_specs}
\small
\begin{tabular}{lccccc}
\toprule
\textbf{Model} & \textbf{Params} & \textbf{Vocab} & \textbf{Hidden} & \textbf{Transformer Blocks} & \textbf{Attention} \\
\midrule
Llama-2-7B  & 6.7B  & 32K  & 4096 & 32 & MHA \\
Llama-2-13B & 13B   & 32K  & 5120 & 40 & MHA \\
Llama-3-8B  & 8B    & 128K & 4096 & 32 & GQA \\
Mistral-7B  & 7.2B  & 32K  & 4096 & 32 & SWA \\
DistilGPT-2 & 0.066B& 50K  & 768  & 12 & MHA \\
\bottomrule
\end{tabular}
\end{table*}

\noindent\textbf{Abbreviations}: MHA = Multi-Head Attention, GQA = Grouped-Query Attention, SWA = Sliding Window Attention.

Llama-2-7B served as the source model for training the RL policy. Llama-2-13B tested zero-shot transfer to a larger model of the same architecture. Llama-3-8B evaluated robustness to denser parameterization and a larger vocabulary. Mistral-7B assessed cross-architecture transfer under sliding-window attention. DistilGPT-2 was used only for prototyping and excluded from final results.

\subsection{Datasets and Calibration}
\label{sec:methodology_datasets}

\subsubsection{Calibration Dataset}

WikiText-2~\citep{merity2017wikitext} (2.4M tokens) is used for calibration. For each model, 128 random sequences ($\sim$20K tokens) are sampled from the training split. Activations are collected via forward passes to compute the 11-dimensional layer embeddings.

\subsubsection{Evaluation Datasets}

Perplexity is measured on the WikiText-2 test split (245K tokens). Downstream zero-shot performance is evaluated on commonsense reasoning benchmarks via \texttt{lm-evaluation-harness}~\citep{gao2021lmevalharness}. The downstream task benchmarks are summarized in \Cref{tab:eval_tasks}.

\begin{table*}[t]
\centering
\caption{Downstream task benchmarks}
\label{tab:eval_tasks}
\begin{small}
\begin{tabular}{lcll}
\toprule
\textbf{Task} & \textbf{Type} & \textbf{Questions} & \textbf{Metric} \\
\midrule
PIQA~\citep{bisk2020piqa}       & Commonsense & 2-choice & Accuracy \\
HellaSwag~\citep{zellers2019hellaswag} & Commonsense & 4-choice & Accuracy \\
WinoGrande~\citep{sakaguchi2020winogrande} & Commonsense & 2-choice & Accuracy \\
ARC~\citep{clark2018arc}        & Reasoning   & Multiple-choice & Accuracy \\
\bottomrule
\end{tabular}
\end{small}
\end{table*}

\subsubsection{Calibration Procedure}

\begin{algorithm}[H]
\caption{Calibration and Embedding Extraction}
\label{alg:calibration}
\begin{algorithmic}[1]
    \State Load pretrained model $M$ in FP16
    \State Sample 128 sequences from WikiText-2 training split
    \For{each layer $i=1$ to $L$}
        \State Collect activations $A_i$ over all sequences
        \State Flatten: $X_i = \text{concatenate}(A_i)$
        \State Compute 11-dimensional embedding $s_i$
    \EndFor
    \State Return $S = [s_1,\dots,s_L]$
\end{algorithmic}
\end{algorithm}

Calibration requires approximately 2 minutes per model.

\subsection{Baseline Methods}
\label{sec:methodology_baselines}

RAMP is compared against FP16 and state-of-the-art 4-bit PTQ methods in in \Cref{tab:baselines_summary}. 

\begin{table}[t]
\centering
\caption{Baseline comparison on Llama-2-7B}
\label{tab:baselines_summary}
\begin{small}
\begin{tabular}{lccc}
\toprule
\textbf{Method} & \textbf{Type} & \textbf{PPL} & \textbf{Size (GB)} \\
\midrule
FP16            & Upper bound   & 5.51 & 13.5 \\
RTN-4           & Naive         & 5.94 & 3.90 \\
GPTQ-4~\citep{frantar2023gptq} & SOTA & 5.69 & 3.90 \\
AWQ-4~\citep{lin2023awq}       & SOTA & 5.60 & 3.90 \\
QUIP\#-4~\citep{tseng2024quip} & SOTA & 5.62 & 3.90 \\
Q4\_K\_M (GGUF) & Deployment    & 5.61 & 3.95 \\
\textbf{RAMP}   & Mixed-precision & \textbf{5.54} & \textbf{3.68} \\
\bottomrule
\end{tabular}
\end{small}
\end{table}

\subsection{Implementation Details and Hyperparameters}
\label{sec:methodology_impl}

SAC hyperparameters are listed in Table~\ref{tab:hyperparams_sac}. Quantization and scale-folding settings appear in Tables~\ref{tab:hyperparams_quant} and \ref{tab:hyperparams_folding}, respectively. Evaluation uses batch size 1, 2048-token context, FP32 precision, greedy decoding (temperature 1.0), and 3 runs (mean $\pm$ std) on NVIDIA RTX PRO 5000 Blackwell and A100 GPUs.

\begin{table*}[t]
\centering
\caption{SAC training hyperparameters}
\label{tab:hyperparams_sac}
\begin{small}
\begin{tabular}{lll}
\toprule
\textbf{Parameter} & \textbf{Value} & \textbf{Rationale} \\
\midrule
Learning rate & $3\times10^{-4}$ & Actor and critics \\
Batch size & 128 & Stability vs.\ efficiency \\
Replay buffer & 30,000 transitions & $\sim$120 episodes $\times$ 250 steps \\
Discount $\gamma$ & 0.99 & Standard episodic RL \\
Target update $\tau$ & 0.005 & Slow tracking \\
Gradient clip & 1.0 & Stability \\
Initial entropy & 0.2 & $\log(0.2)\approx-1.6$ \\
Target entropy & $-d_a$ & Automatic \\
Networks & [512,512,256] (4 layers) & Capacity with tractable cost \\
Activation & ReLU & Fast inference \\
Normalization & LayerNorm & Per hidden layer \\
Optimizer & Adam & Adaptive \\
Warm-up episodes & 20 & Random exploration \\
Max episodes & 200--250 & Empirical convergence \\
\bottomrule
\end{tabular}
\end{small}
\end{table*}

\begin{table*}[t]
\centering
\caption{Quantization configuration}
\label{tab:hyperparams_quant}
\begin{small}
\begin{tabular}{lll}
\toprule
\textbf{Parameter} & \textbf{Value} & \textbf{Rationale} \\
\midrule
Method & Per-group asymmetric & Standard LLM PTQ \\
Group size & 128 & GGUF-compatible \\
Bit widths & $\{3,4,5\}$ & Aggressive compression range \\
Target avg.\ bits & 4.0 & Compression goal \\
Ceiling & 4.25 & Reward-enforced \\
Calibration & WikiText-2 (128 seq., 2048 tokens) & Efficient and representative \\
Scale folding & Enabled & Sub-4-bit stability \\
\bottomrule
\end{tabular}
\end{small}
\end{table*}

\begin{table*}[t]
\centering
\caption{Scale folding configuration}
\label{tab:hyperparams_folding}
\begin{small}
\begin{tabular}{lll}
\toprule
\textbf{Parameter} & \textbf{Value} & \textbf{Rationale} \\
\midrule
Scale & $s=\sqrt{\text{act\_scale}}$ & Smoothing \\
Normalization & $s\leftarrow s/\text{mean}(s)$ & Magnitude preservation \\
Stabilization & $+10^{-5}$ & Numerical safety \\
Scope & Per-channel (Q/K/V/O/gate/up) & Fine-grained conditioning \\
Norm adjustment & $\text{RMSNorm}\leftarrow\text{RMSNorm}/s$ & Equivalence preservation \\
Calibration & 128 sequences & Reuse of quantization data \\
\bottomrule
\end{tabular}
\end{small}
\end{table*}

\subsection{Pareto Frontier Analysis}
\label{sec:results_pareto}

\subsubsection{Llama-2-7B}

Quantitative results are presented in \Cref{tab:results_llama2_7b}.

\begin{table*}[t]
\centering
\caption{Results on Llama-2-7B (224 quantizable linear layers)}
\label{tab:results_llama2_7b}
\begin{small}
\begin{tabular}{lcccc}
\toprule
\textbf{Method} & \textbf{Size (GB)} & \textbf{PPL} & \textbf{Size vs RAMP} & \textbf{PPL vs RAMP} \\
\midrule
FP16 & 13.5 & 5.51 & $+267\%$ & — \\
RTN-4 & 3.90 & 5.94 & $+6.0\%$ & $+7.2\%$ \\
GPTQ-4 & 3.90 & 5.69 & $+6.0\%$ & $+2.7\%$ \\
AWQ-4 & 3.90 & 5.60 & $+6.0\%$ & $+1.1\%$ \\
QUIP\#-4 & 3.90 & 5.62 & $+6.0\%$ & $+1.4\%$ \\
Q4\_K\_M & 3.95 & 5.61 & $+7.3\%$ & $+1.3\%$ \\
\textbf{RAMP} & \textbf{3.68} & \textbf{5.54} & — & — \\
\bottomrule
\end{tabular}
\end{small}
\end{table*}

\begin{figure}[t]
    \centering
    \includegraphics[width=\columnwidth]{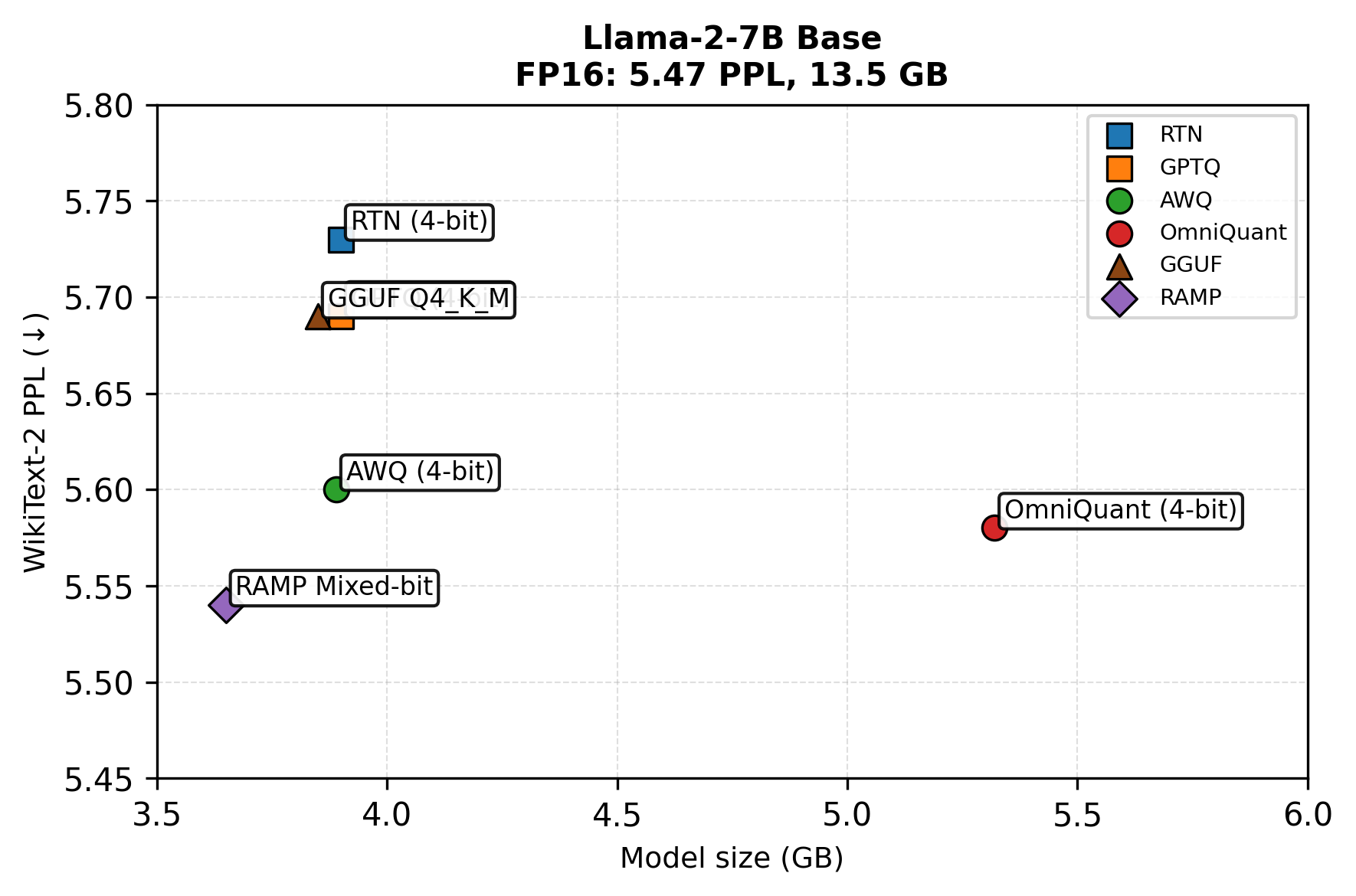}
    \caption{Perplexity vs.\ model size on Llama-2-7B. RAMP dominates uniform 4-bit baselines.}
    \label{fig:llama2_7b_tradeoff}
\end{figure}

\begin{figure}[t]
    \centering
    \includegraphics[width=\columnwidth]{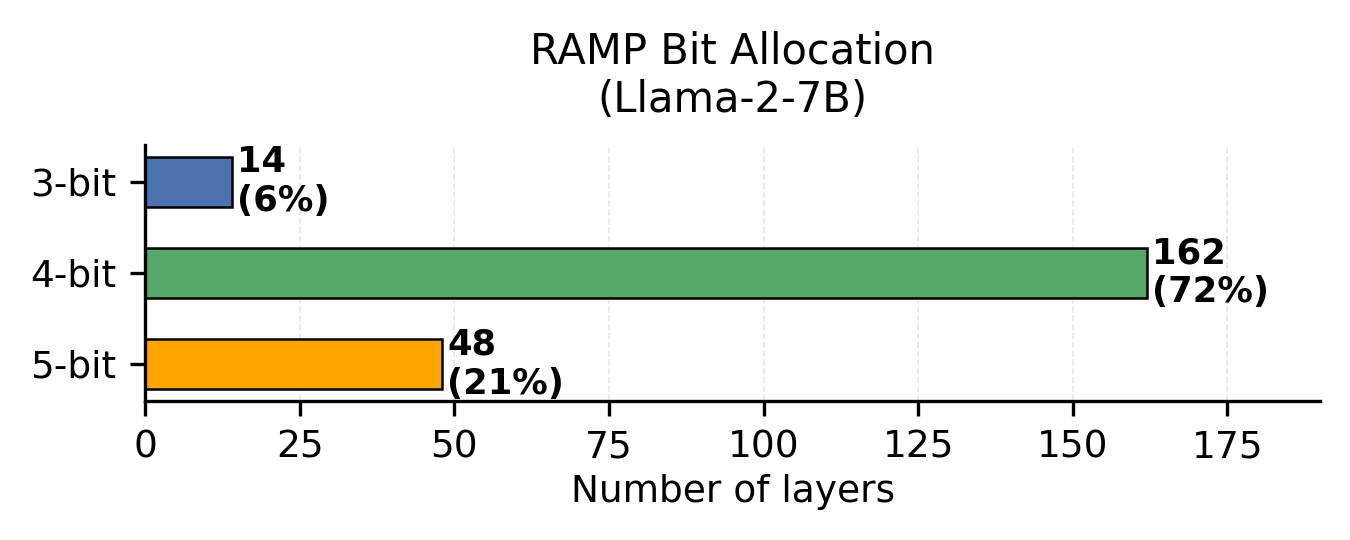}
    \caption{RAMP bit allocation on Llama-2-7B (224 quantizable linear layers).}
    \label{fig:ramp_bit_allocation_llama2}
\end{figure}

RAMP is strictly smaller and higher-quality than all 4-bit baselines, establishing a Pareto frontier. The Pareto frontier is visualized in \Cref{fig:llama2_7b_tradeoff}, and the concrete bit allocation is shown in \Cref{fig:ramp_bit_allocation_llama2}.

\subsubsection{Llama-2-13B (Zero-Shot Transfer)}

Results on Llama-2-13B are reported in \Cref{tab:results_llama2_13b}.

\begin{table*}[t]
\centering
\caption{Results on Llama-2-13B (280 quantizable linear layers)}
\label{tab:results_llama2_13b}
\begin{small}
\begin{tabular}{lcccc}
\toprule
\textbf{Method} & \textbf{Size (GB)} & \textbf{PPL} & \textbf{Size vs RAMP} & \textbf{PPL vs RAMP} \\
\midrule
FP16 & 26.0 & 4.85 & $+259\%$ & — \\
GPTQ-4 & 7.68 & 5.02 & $+6.1\%$ & $+1.4\%$ \\
AWQ-4 & 7.68 & 4.97 & $+6.1\%$ & $+0.4\%$ \\
Q4\_K\_M & 7.73 & 4.98 & $+6.8\%$ & $+0.6\%$ \\
Direct SAC & 7.25 & 4.96 & $+0.1\%$ & $+0.2\%$ \\
\textbf{RAMP (zero-shot)} & \textbf{7.24} & \textbf{4.95} & — & — \\
\bottomrule
\end{tabular}
\end{small}
\end{table*}

Zero-shot transfer from Llama-2-7B marginally outperforms direct training on the target model.

\subsubsection{Llama-3-8B}

Results on Llama-3-8B appear in \Cref{tab:results_llama3_8b}.

\begin{table*}[t]
\centering
\caption{Results on Llama-3-8B}
\label{tab:results_llama3_8b}
\begin{small}
\begin{tabular}{lcccc}
\toprule
\textbf{Method} & \textbf{Size (GB)} & \textbf{PPL} & \textbf{Size vs RAMP} & \textbf{PPL vs RAMP} \\
\midrule
FP16 & 16.1 & 6.23 & $+281\%$ & — \\
GPTQ-4 & 5.74 & 8.58 & $+36\%$ & $+33\%$ \\
AWQ-4 & 5.31 & 6.74 & $+26\%$ & $+4.2\%$ \\
Q4\_K\_M & 5.48 & 6.79 & $+30\%$ & $+5.0\%$ \\
\textbf{RAMP (zero-shot)} & \textbf{4.22} & \textbf{6.47} & — & — \\
\bottomrule
\end{tabular}
\end{small}
\end{table*}

\begin{figure}[t]
    \centering
    \includegraphics[width=\columnwidth]{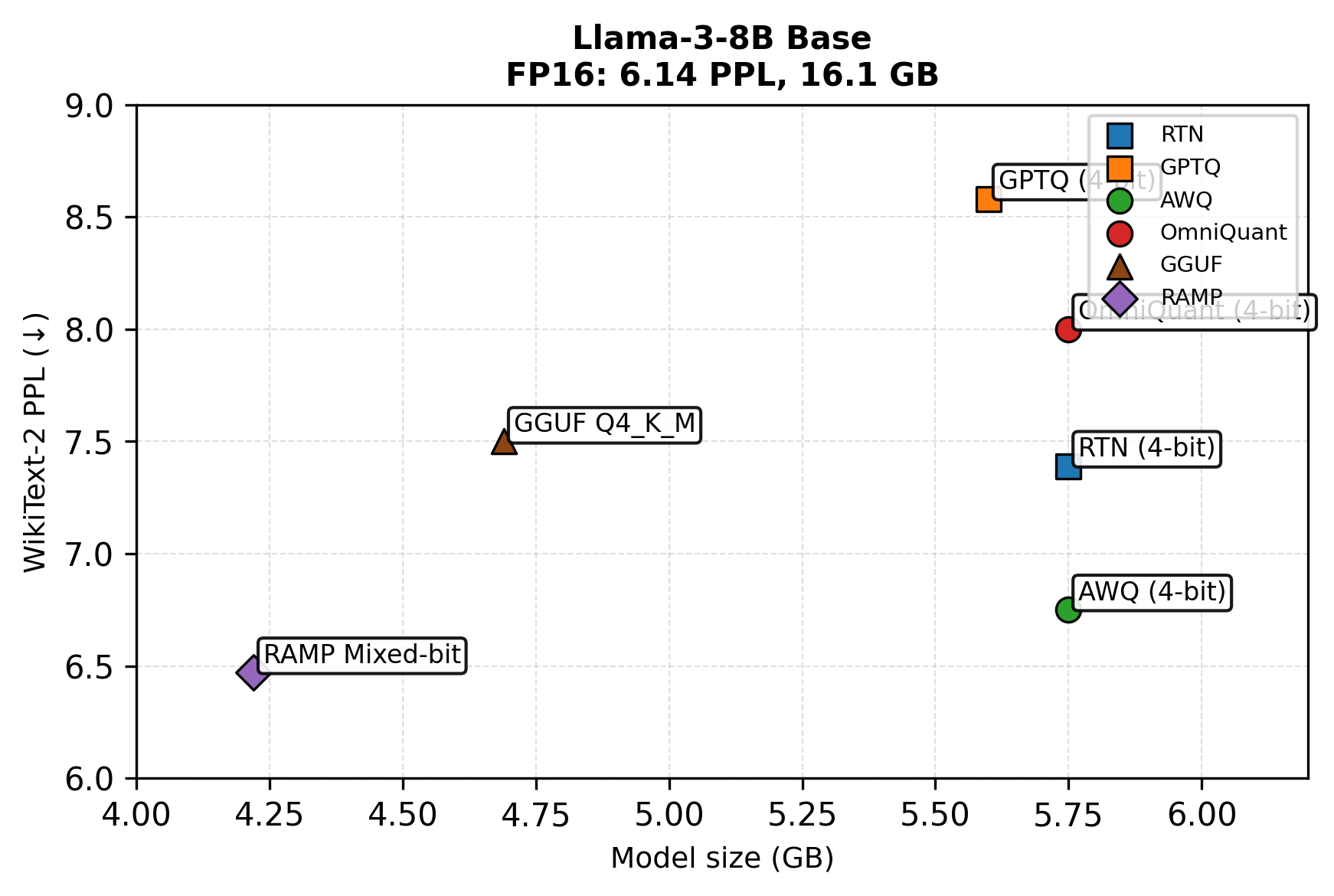}
    \caption{Perplexity vs.\ model size on Llama-3-8B.}
    \label{fig:llama3_8b_tradeoff}
\end{figure}

\begin{figure}[t]
    \centering
    \includegraphics[width=\columnwidth]{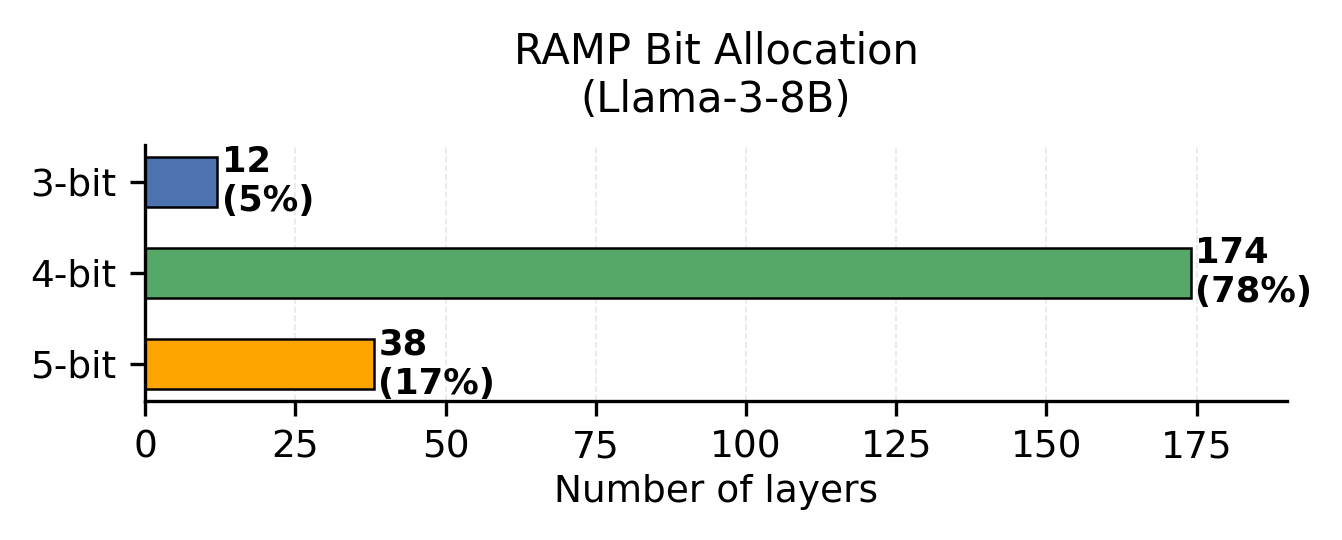}
    \caption{RAMP bit allocation on Llama-3-8B (224 quantizable linear layers).}
    \label{fig:ramp_bit_allocation_llama3}
\end{figure}

RAMP remains effective despite denser parameterization and larger vocabulary.
The trade-off is shown in \Cref{fig:llama3_8b_tradeoff}, with the bit allocation in \Cref{fig:ramp_bit_allocation_llama3}.

\subsubsection{Mistral-7B (Cross-Architecture Transfer)}

Results on Mistral-7B are shown in \Cref{tab:results_mistral_7b}.

\begin{table*}[t]
\centering
\caption{Results on Mistral-7B}
\label{tab:results_mistral_7b}
\begin{small}
\begin{tabular}{lcccc}
\toprule
\textbf{Method} & \textbf{Size (GB)} & \textbf{PPL} & \textbf{Size vs RAMP} & \textbf{PPL vs RAMP} \\
\midrule
FP16 & 14.5 & 5.45 & $+299\%$ & — \\
GPTQ-4 & 3.85 & 5.71 & $+6.1\%$ & $+2.7\%$ \\
AWQ-4 & 3.85 & 5.59 & $+6.1\%$ & $+0.5\%$ \\
Q4\_K\_M & 3.90 & 5.60 & $+7.4\%$ & $+0.7\%$ \\
Direct SAC & 3.62 & 5.58 & $-0.3\%$ & $+0.4\%$ \\
\textbf{RAMP (zero-shot)} & \textbf{3.63} & \textbf{5.56} & — & — \\
\bottomrule
\end{tabular}
\end{small}
\end{table*}

Zero-shot transfer again matches or exceeds direct training.

\begin{figure}[t]
    \centering
    \includegraphics[width=\columnwidth]{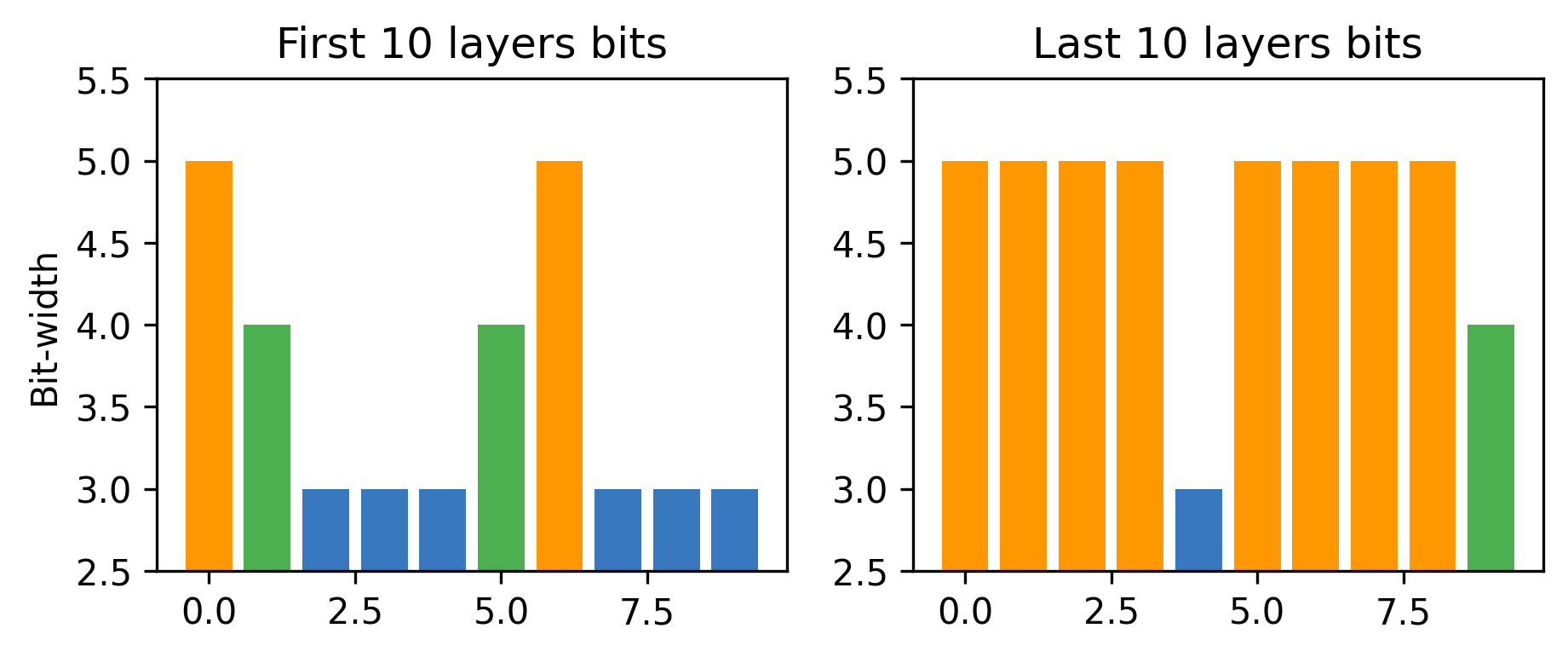}
    \caption{First- and last-10 layer bit assignments across models.}
    \label{fig:ramp_first_last_layers}
\end{figure}

\subsection{Zero-Shot Policy Transfer}
\label{sec:results_transfer}

Zero-shot vs. direct training comparison is provided in \Cref{tab:results_transfer_summary}.

\begin{table*}[t]
\centering
\caption{Zero-shot transfer summary}
\label{tab:results_transfer_summary}
\small
\begin{tabular}{lccccc}
\toprule
\textbf{Target} & \textbf{Direct PPL} & \textbf{Direct Bits} & \textbf{Zero-shot PPL} & \textbf{Zero-shot Bits} & \textbf{Winner} \\
\midrule
Llama-2-13B & 4.96 & 3.66 & 4.95 & 3.66 & Zero-shot \\
Mistral-7B  & 5.58 & 3.67 & 5.56 & 3.67 & Zero-shot \\
\bottomrule
\end{tabular}
\end{table*}

Zero-shot policies trained on Llama-2-7B consistently outperform or match direct training on target models, confirming that the 11-dimensional embeddings capture architecture-level sensitivity.

Layer-wise bit patterns exhibit strong depth-wise correlation ($\rho\approx0.9$) across models, with higher precision at input/output layers and aggressive compression in intermediate layers.

First- and last-10 layer bit assignments across models are illustrated in \Cref{fig:ramp_first_last_layers}.

\subsection{Downstream Task Performance}
\label{sec:results_downstream}

Commonsense reasoning performance is reported in \Cref{tab:results_downstream}.

\begin{table*}[t]
\centering
\caption{Commonsense reasoning performance (Llama-2-7B)}
\label{tab:results_downstream}
\begin{small}
\begin{tabular}{lccccc}
\toprule
\textbf{Method} & \textbf{PIQA} & \textbf{HellaSwag} & \textbf{WinoGrande} & \textbf{ARC} & \textbf{Avg} \\
\midrule
FP16 & 79.4\% & 58.2\% & 64.0\% & 53.9\% & 63.9\% \\
AWQ-4 & 78.8\% & 57.1\% & 62.3\% & 52.4\% & 62.7\% \\
Q4\_K\_M & 78.6\% & 56.9\% & 62.1\% & 52.1\% & 62.4\% \\
\textbf{RAMP} & \textbf{79.2\%} & \textbf{57.9\%} & \textbf{63.7\%} & \textbf{53.6\%} & \textbf{63.6\%} \\
\midrule
Retention vs.\ FP16 & 99.7\% & 99.5\% & 99.5\% & 99.4\% & \textbf{99.5\%} \\
\bottomrule
\end{tabular}
\end{small}
\end{table*}

RAMP retains 99.5\% of FP16 downstream accuracy on average while achieving substantially higher compression than uniform 4-bit baselines.

\section{Analysis \& Discussion}
\label{sec:analysis}

\subsection{Why Zero-Shot Transfer Works}
\label{sec:analysis_transfer}

Policies trained on Llama-2-7B consistently match or outperform those trained directly on target models (Llama-2-13B, Mistral-7B), suggesting the learned policy captures architecture-level rather than instance-specific patterns.

\subsubsection{Embedding Stability Across Models}

Embedding stability across models is quantified in \Cref{tab:embedding_stability}.

\begin{table*}[t]
\centering
\caption{Embedding stability: Pearson correlations across models}
\label{tab:embedding_stability}
\begin{small}
\begin{tabular}{lcccc}
\toprule
\textbf{Layer Role} & \textbf{Llama-2-7B} & \textbf{Llama-2-13B} & \textbf{Mistral-7B} & \textbf{Correlation} \\
\midrule
Embedding layer     & (0.87, 0.34, 0.12, ...) & (0.91, 0.36, 0.13, ...) & (0.85, 0.32, 0.11, ...) & $r=0.94$ \\
Attention o\_proj   & (3.12, 0.18, 0.08, ...) & (3.15, 0.19, 0.09, ...) & (3.08, 0.17, 0.07, ...) & $r=0.96$ \\
MLP down\_proj      & (2.87, 0.22, 0.11, ...) & (2.91, 0.23, 0.12, ...) & (2.84, 0.21, 0.10, ...) & $r=0.95$ \\
Output layer        & (1.24, 0.41, 0.19, ...) & (1.28, 0.42, 0.20, ...) & (1.21, 0.40, 0.18, ...) & $r=0.93$ \\
\bottomrule
\end{tabular}
\end{small}
\end{table*}

Correlations $r > 0.93$ confirm that normalized 11-dimensional embeddings reliably encode structural role independent of model scale or exact parameter values.

\subsubsection{Overfitting in Direct Training}

Direct training on larger or architecturally distinct models risks overfitting to noise in layer-specific dynamics, initialization effects, and gradient fluctuations. Larger models (more layers, longer episodes) exacerbate this due to increased state-space complexity and fewer effective updates within fixed wall-clock time. Evidence: zero-shot transfer from Llama-2-7B (32 blocks / 224 layers) yields 4.95 PPL on Llama-2-13B (40 blocks / 280 layers) vs.\ 4.96 PPL from direct training.

\subsubsection{Policy Regularization}

The zero-shot policy, trained on the simpler 32-block Llama-2-7B model, produces more stable bit allocations when applied to larger or architecturally distinct targets. This is consistent with learning generalizable patterns that avoid overfitting to model-specific noise or initialization effects.

\subsection{Structural Sensitivity Patterns}
\label{sec:analysis_patterns}

The learned policy reveals consistent depth-wise and type-specific trends.

\subsubsection{Depth-wise Allocation}

Precision increases toward the output: early layers average $\sim$3.77 bits (mixed 3--5), middle layers $\sim$4.01 bits (mostly 3--4), late layers $\sim$4.52 bits (enriched for 5-bit).

\begin{figure}[t]
    \centering
    \includegraphics[width=\columnwidth]{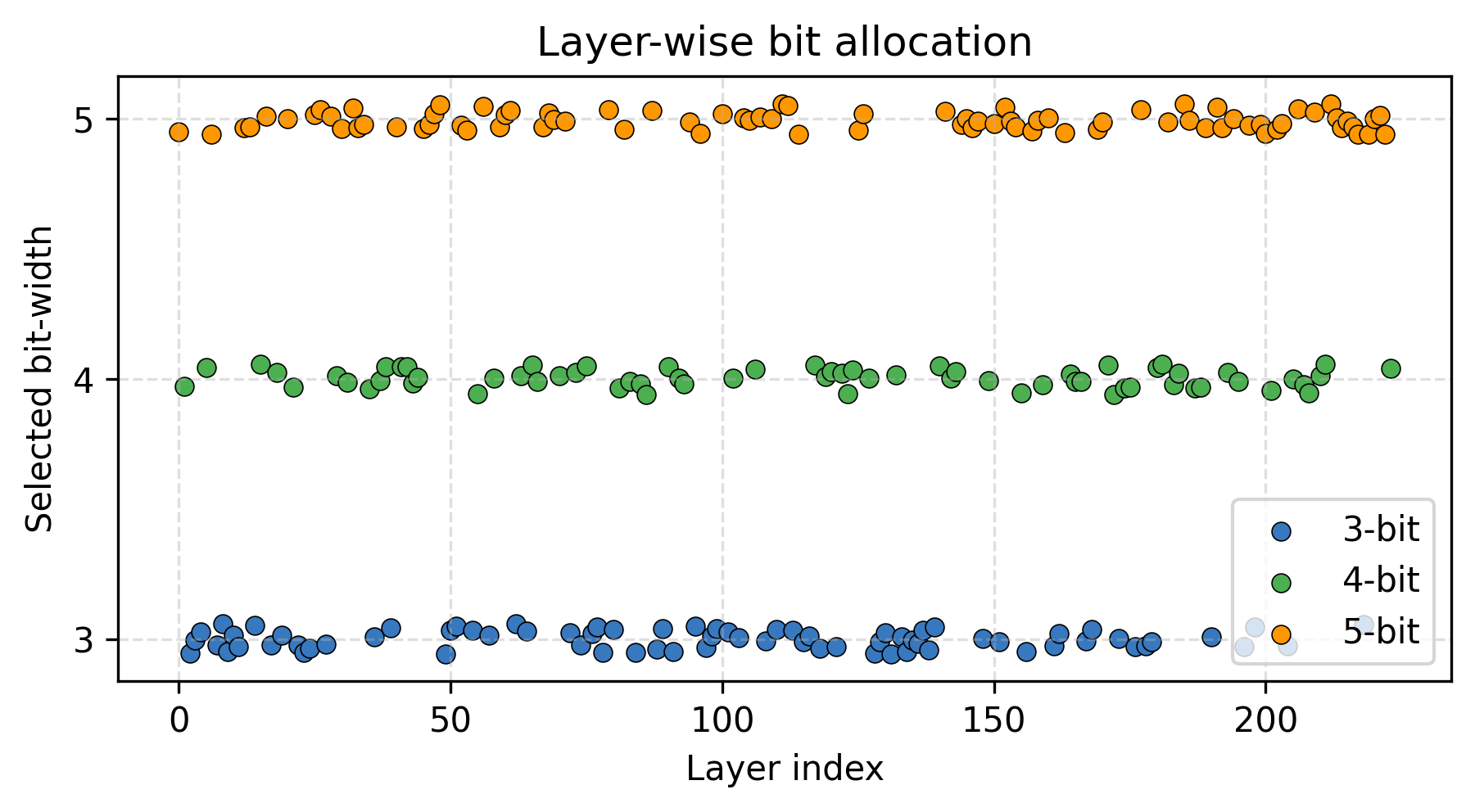}
    \caption{Layer-wise bit assignments vs.\ depth (224 quantizable linear layers).}
    \label{fig:ramp_layerwise_bits}
\end{figure}

\begin{figure}[t]
    \centering
    \includegraphics[width=\columnwidth]{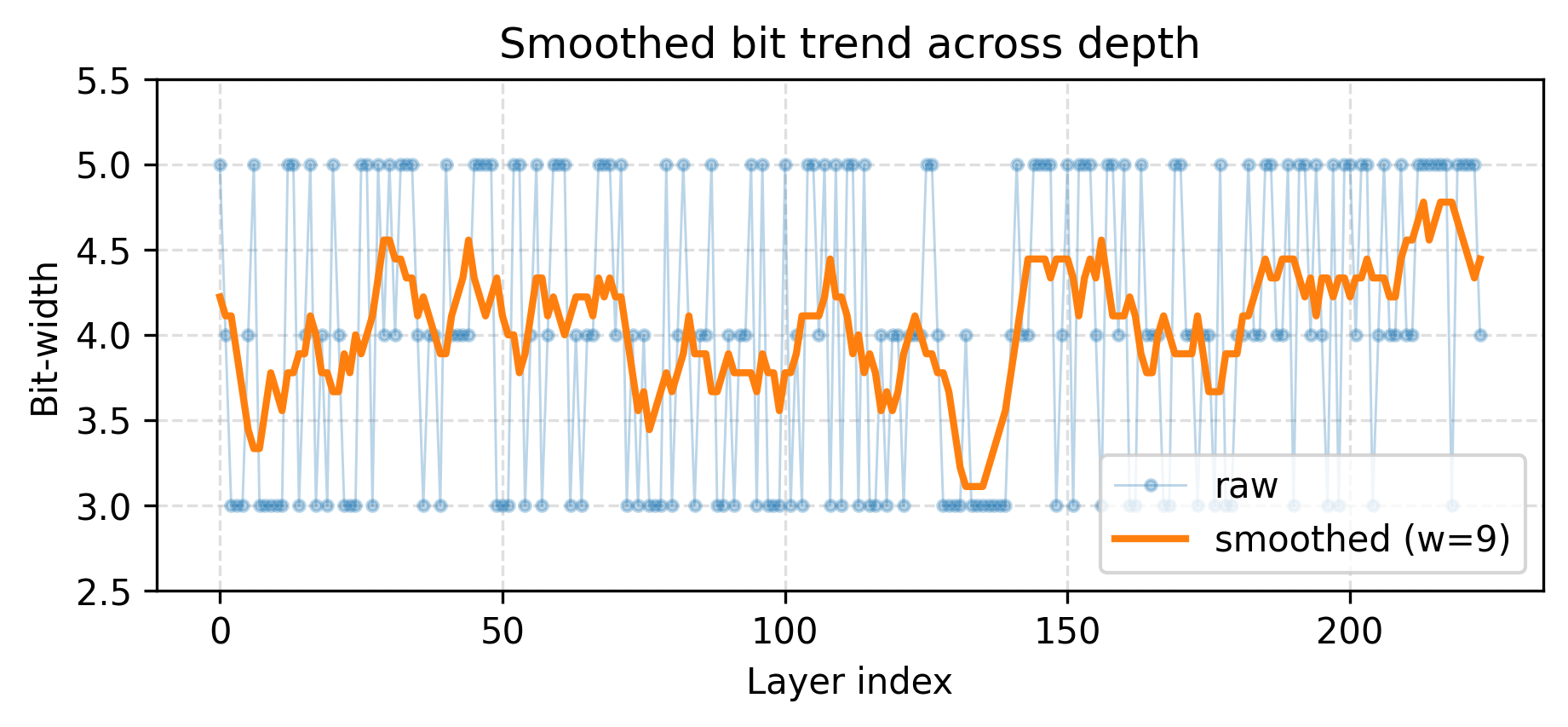}
    \caption{Smoothed bit-width trend across depth.}
    \label{fig:ramp_smoothed_bit_trend}
\end{figure}

This asymmetric pattern prioritizes output projections and logits while aggressively compressing redundant intermediate layers.

\begin{figure}[t]
    \centering
    \includegraphics[width=\columnwidth]{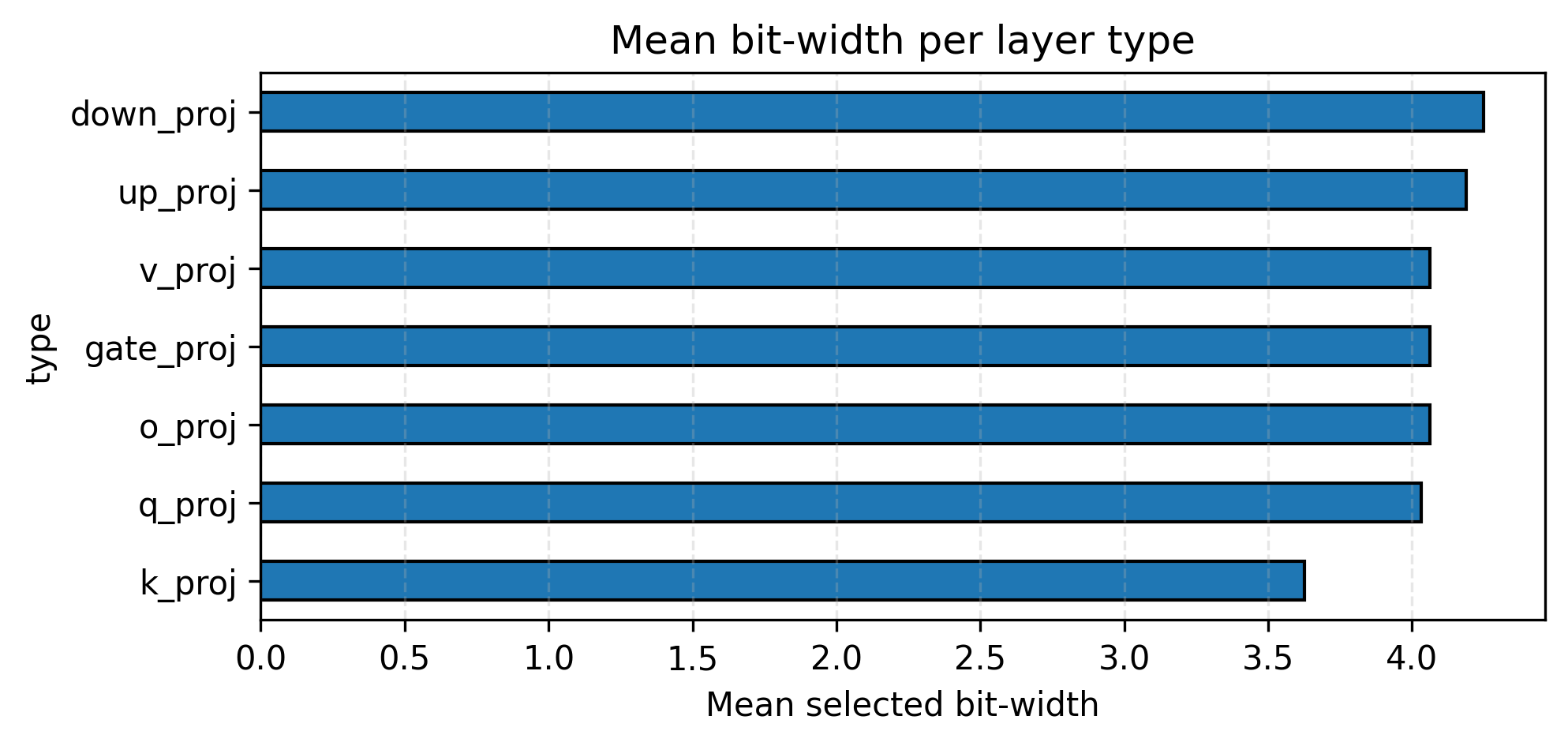}
    \caption{Mean bit-width per layer type.}
    \label{fig:ramp_mean_bit_per_type}
\end{figure}

Layer-wise bit assignments versus depth are shown in \Cref{fig:ramp_layerwise_bits} and the smoothed trend in \Cref{fig:ramp_smoothed_bit_trend}. Mean bit-width per layer type is displayed in \Cref{fig:ramp_mean_bit_per_type}.

\subsubsection{Activation Outliers and Bit Choices}

Simple statistics like $\log(\text{Act\_Max})$ or activation-importance scores show weak correlation with assigned bits ($|r| \approx 0.03$). The policy relies on the full compositional 11D embedding rather than any single outlier proxy.

\subsubsection{Exploration Dynamics}

\begin{figure}[t]
    \centering
    \includegraphics[width=\columnwidth]{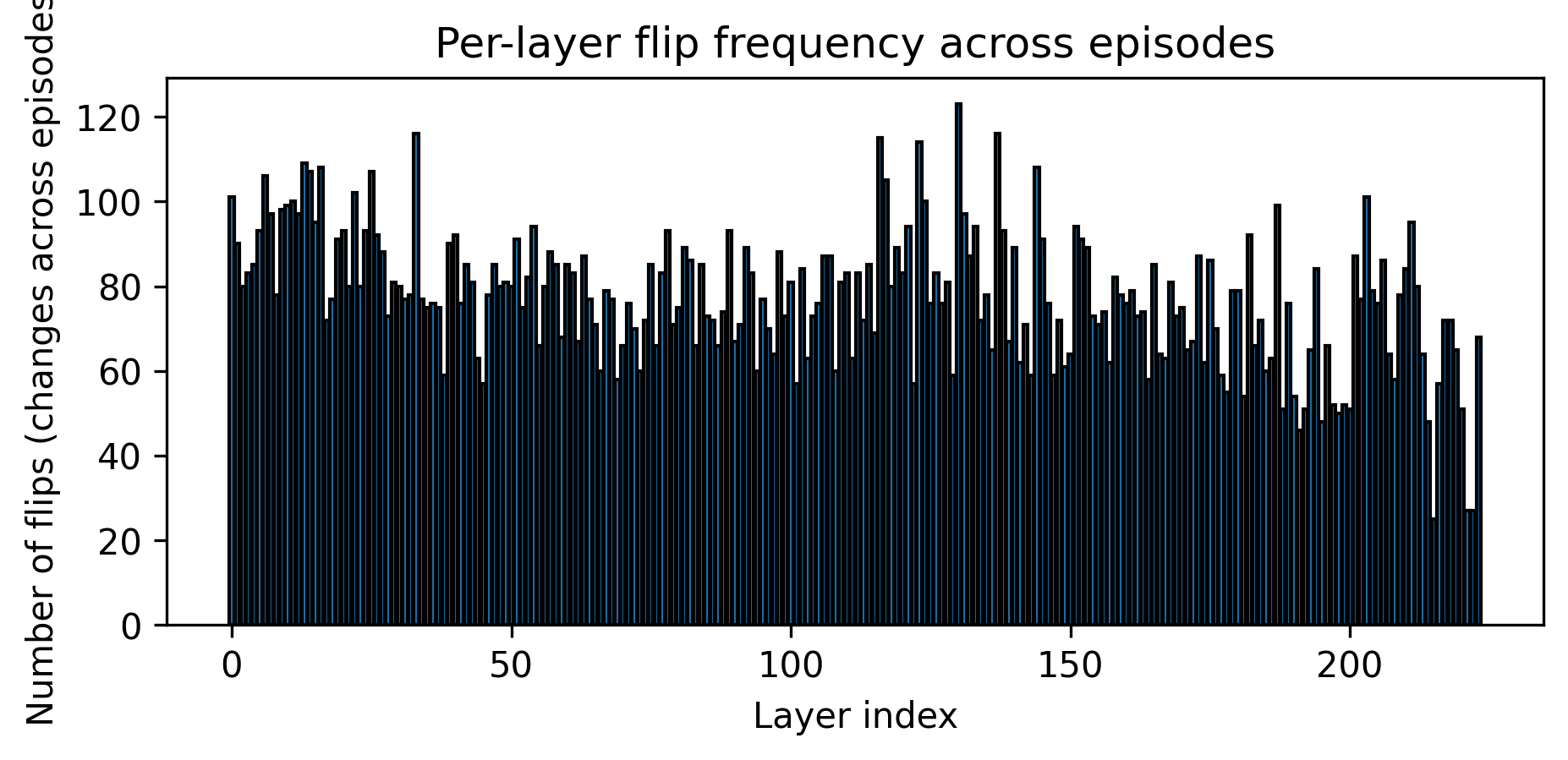}
    \caption{Per-layer bit-flip frequency during training.}
    \label{fig:ramp_flip_frequency}
\end{figure}

\begin{figure}[t]
    \centering
    \includegraphics[width=\columnwidth]{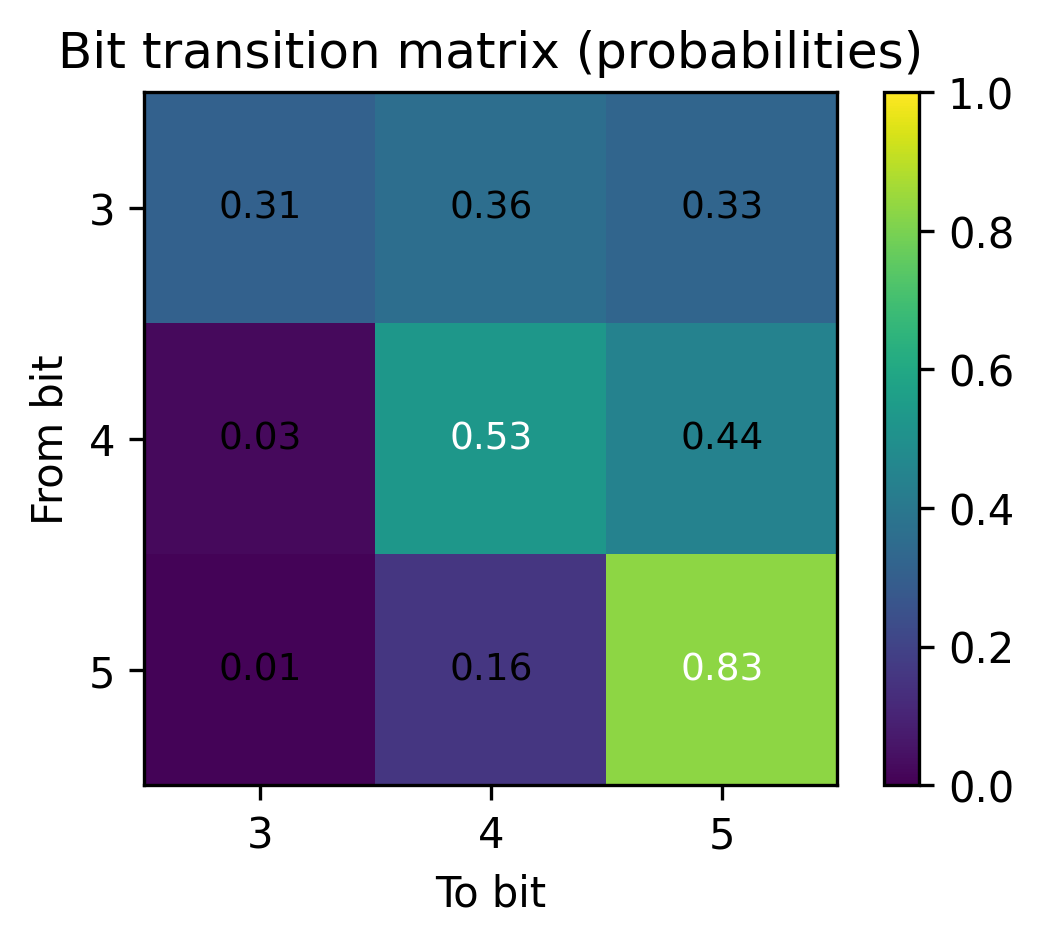}
    \caption{Bit-transition probabilities across episodes.}
    \label{fig:ramp_transition_matrix}
\end{figure}

Flip frequency and transition matrix reveal stable convergence: most layers settle early, with exploration concentrated on boundary-sensitive layers.

Per-layer bit-flip frequency during training is analyzed in \Cref{fig:ramp_flip_frequency}, and transition probabilities are shown in \Cref{fig:ramp_transition_matrix}.

\subsubsection{Allocation by Layer Type}

Output projections and final layers receive highest average precision; intermediate MLP down-projections are most aggressively compressed.

\subsection{Ablation Studies}
\label{sec:analysis_ablation}

\subsubsection{Scale Folding}

Scale folding impact is quantified in \Cref{tab:ablation_folding} and \Cref{tab:ablation_sf}.

\begin{table*}[t]
\centering
\caption{Scale folding ablation}
\label{tab:ablation_sf}
\begin{small}
\begin{tabular}{lccc}
\toprule
\textbf{Configuration} & \textbf{Episodes to Conv.} & \textbf{Final PPL} & \textbf{Stable?} \\
\midrule
No folding & $>250$ & 5.58 & Partial \\
With folding & 150 & 5.54 & $\checkmark$ \\
\bottomrule
\end{tabular}
\end{small}
\end{table*}

Scale folding is required for stable convergence and low-bit viability.

\subsubsection{Reward Design}

Reward design ablation is shown in \Cref{tab:ablation_reward}.

\begin{table*}[t]
\centering
\caption{Reward ablation}
\label{tab:ablation_reward}
\begin{small}
\begin{tabular}{lccc}
\toprule
\textbf{Reward} & \textbf{Episodes to Conv.} & \textbf{Final PPL} & \textbf{Variance} \\
\midrule
Naive (Acc - $\lambda$ Bits) & $\infty$ & $>5.75$ & 0.35 \\
Linear & 200 & 5.61 & 0.18 \\
Quality-first (asymmetric + cliff) & 150 & 5.54 & 0.05 \\
\bottomrule
\end{tabular}
\end{small}
\end{table*}

Asymmetric quality prioritization with cliff penalty yields 33\% faster convergence and 7$\times$ lower variance.

\subsubsection{SAC vs.\ PPO}

Comparison of SAC versus PPO is provided in \Cref{tab:ablation_sac_vs_ppo}.

\begin{table*}[t]
\centering
\caption{SAC vs.\ PPO}
\label{tab:ablation_sac_vs_ppo}
\begin{small}
\begin{tabular}{lccc}
\toprule
\textbf{Algorithm} & \textbf{GPU Hours} & \textbf{Final PPL} & \textbf{Variance} \\
\midrule
PPO & 48 & 5.62 & 0.15 \\
SAC & 6 & 5.54 & 0.03 \\
\bottomrule
\end{tabular}
\end{small}
\end{table*}

SAC is 8$\times$ more sample-efficient due to replay buffer reuse, producing lower perplexity and greater stability.

\subsubsection{Embedding Expressivity}

\begin{figure}[t]
    \centering
    \includegraphics[width=\columnwidth]{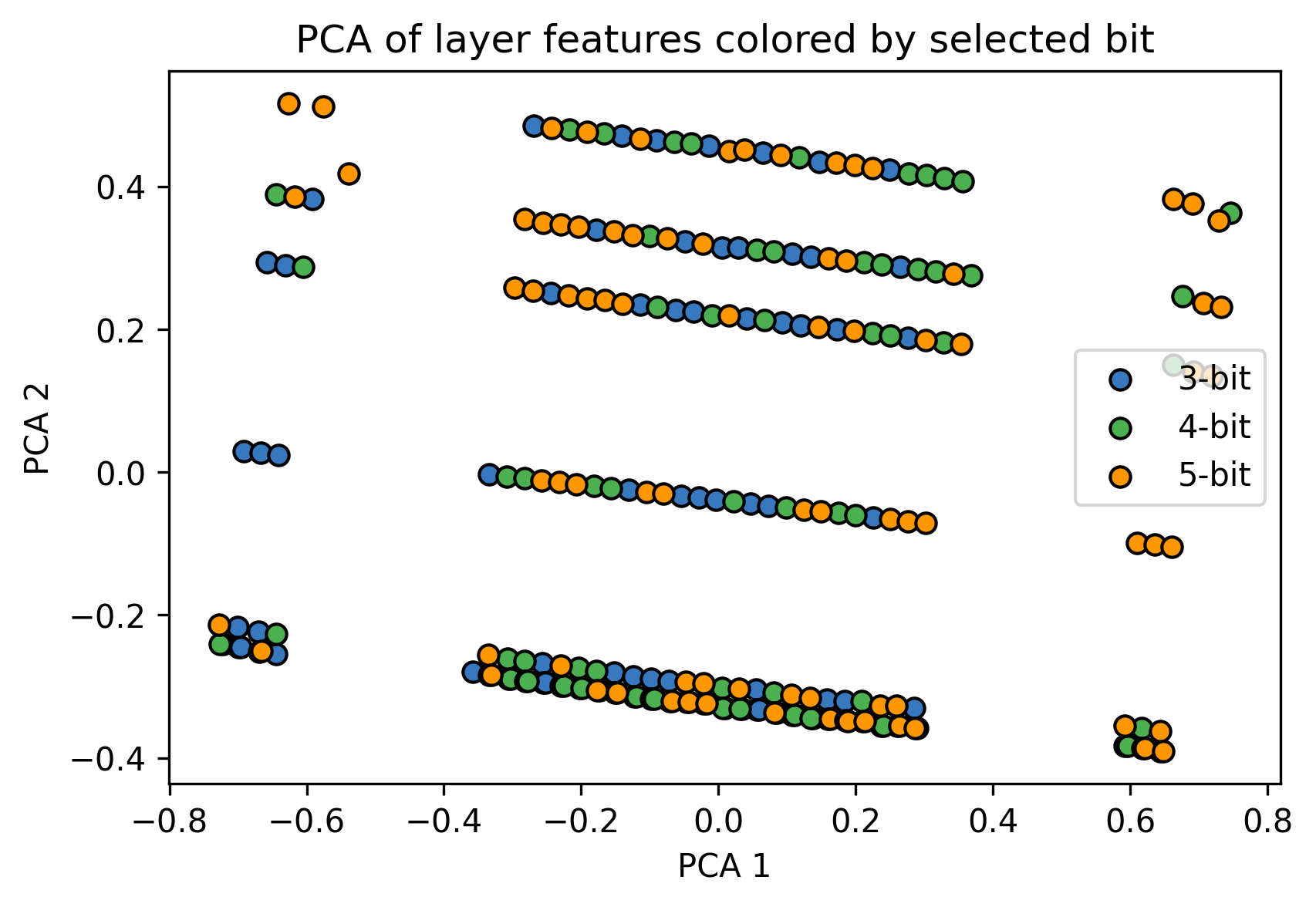}
    \caption{PCA of layer embeddings colored by assigned bit-width.}
    \label{fig:ramp_layer_features_pca}
\end{figure}

The 11D embedding captures compositional sensitivity beyond single features ($|r| \leq 0.30$ for individual dimensions).The PCA projection of layer embeddings colored by assigned bit-width is presented in \Cref{fig:ramp_layer_features_pca}.

\subsubsection{HALO Deployment Trade-offs}

HALO deployment trade-offs compared to custom kernels are summarized in \Cref{tab:ablation_halo}.

\begin{table*}[t]
\centering
\caption{HALO vs.\ custom kernels}
\label{tab:ablation_halo}
\begin{small}
\begin{tabular}{lcc}
\toprule
\textbf{Approach} & \textbf{Throughput} & \textbf{Portability / Effort} \\
\midrule
Custom mixed-precision kernels & Highest (hardware-specific) & Low (high engineering) \\
HALO / GGUF & Competitive & High (zero custom dev, broad hardware) \\
\bottomrule
\end{tabular}
\end{small}
\end{table*}

HALO trades marginal peak throughput for mature, portable, community-optimized kernels across CPU/GPU/Apple Silicon without custom development.

\subsection{Limitations and Future Directions}
\label{sec:analysis_limitations}

\subsubsection{Current Limitations}

RAMP was evaluated exclusively on decoder-only Transformer architectures (Llama and Mistral families). Its applicability to encoder-decoder models (e.g., T5, mT5) and mixture-of-experts architectures remains untested. The method currently supports only discrete bit-widths $\{3,4,5,6\}$ compatible with existing kernels; fractional or mixed-precision formats (e.g., 2.5-bit, 3.5-bit) are unsupported. Quantization occurs at layer granularity; finer head-level or channel-level allocation could yield further gains but requires corresponding kernel support. RAMP operates in the post-training setting only; integration with quantization-aware training could recover additional accuracy at aggressive compression levels. Although the policy transfers zero-shot, calibration (embedding extraction) must be repeated per model. Finally, bit allocations are static; input-dependent dynamic quantization is left for future work.

\subsubsection{Future Directions}

Potential extensions include:
\begin{itemize}
    \item Cross-architecture transfer to encoder-decoder and non-Transformer models, requiring architecture-agnostic state representations.
    \item Head- or channel-level mixed precision, contingent on kernel support for intra-layer variable bit-widths.
    \item Integration with quantization-aware training to reach sub-3-bit regimes.
    \item Input-adaptive dynamic policies that adjust bits at inference time.
    \item Multi-objective optimization jointly targeting perplexity, latency, power, and size.
    \item Hardware co-design of kernels optimized for RAMP's typical bit distributions.
    \item Hybrid sparsity-plus-quantization pipelines.
    \item Continual adaptation of policies for incrementally trained or fine-tuned models.
\end{itemize}

\section{Conclusion}
\label{sec:conclusion}

RAMP learns transferable mixed-precision quantization policies via reinforcement learning, achieving Pareto-optimal accuracy-efficiency trade-offs while enabling deployment through standardized GGUF formats.

\subsection{Key Contributions}

The main contributions are:
\begin{enumerate}
    \item Demonstration of the first transferable quantization policy for LLMs: a policy trained on Llama-2-7B generalizes zero-shot to Mistral-7B and Llama-2-13B, often yielding lower perplexity than policies trained directly on the target.
    \item Scale Folding, a preconditioning technique that stabilizes activation distributions and enables reliable sub-4-bit quantization.
    \item A quality-prioritized reward with asymmetric penalties and cliff constraints that drives rapid, stable convergence of the SAC agent.
    \item Strictly superior Pareto frontiers: on Llama-2-7B, RAMP reaches 5.54 PPL at 3.68\,GB versus AWQ's 5.60 PPL at 3.90\,GB (6\% smaller and 1.1\% higher quality); comparable gains hold across Llama-3-8B and Mistral-7B.
    \item HALO, a hardware-aware export pipeline that produces GGUF models runnable on CPUs, GPUs, Apple Silicon, and select edge devices without custom kernels.
\end{enumerate}

\subsection{Outlook}

RAMP establishes that quantization sensitivity is largely a structural property of Transformer architectures rather than model-instance specific. This insight suggests that reinforcement-learned policies may generalize to other compression techniques (pruning, sparsity, distillation) and opens a path toward general quantization agents that map any model and efficiency target to an optimal quantized artifact.

\section{Broader Context and Impact}
\label{sec:context}

RAMP enables practical mixed-precision deployment on consumer and edge hardware via standardized \texttt{llama.cpp} backends. A single GGUF artifact runs unmodified across NVIDIA/AMD GPUs, x86 CPUs, Apple Silicon, and some mobile platforms, reducing the barrier between research prototypes and real-world inference.

Model compression also lowers energy consumption. Illustrative estimates for 8\,h/day inference of a 7B-class model show that RAMP-quantized execution on consumer hardware can reduce annual carbon emissions by 66--75\% relative to cloud A100 baselines (see Table~\ref{tab:sustainability} for assumptions and order-of-magnitude calculations).

\begin{table*}[t]
\centering
\caption{Illustrative annual energy and carbon footprint (7B model, 8\,h/day)}
\label{tab:sustainability}
\begin{small}
\begin{tabular}{lcccc}
\toprule
\textbf{Deployment} & \textbf{Power (W)} & \textbf{Annual kWh} & \textbf{CO$_2$ (kg)} & \textbf{Cost (\$)} \\
\midrule
Cloud A100 & 350 & 2{,}920 & 1{,}168 & 292 \\
RTX 3090 FP16 & 250 & 2{,}088 & 835 & 209 \\
\textbf{RTX 3090 RAMP} & \textbf{85} & \textbf{711} & \textbf{284} & \textbf{71} \\
Xeon CPU RAMP & 65 & 546 & 218 & 55 \\
M1 laptop RAMP & 12 & 101 & 40 & 10 \\
\bottomrule
\end{tabular}
\end{small}
\end{table*}

\footnotesize
Assumptions: US grid intensity 0.4\,kg CO$_2$/kWh, \$0.10/kWh cloud pricing, public TDP and benchmark power draws. These are order-of-magnitude estimates, not direct measurements.

\subsection{Future of Efficient LLM Deployment}

Further gains are expected from hybrid sparsity-plus-quantization, input-adaptive policies, and hardware co-design of mixed-precision kernels. The long-term goal is a general quantization agent capable of mapping arbitrary models and efficiency targets to optimal compressed artifacts across diverse hardware.

\FloatBarrier
\bibliographystyle{plainnat}
\bibliography{ramp_report_references}

@article{frantar2023gptq,
  title={GPTQ: Accurate Post-Training Quantization for Generative Pre-trained Transformers},
  author={Frantar, Elias and Ashkboos, Saleh and Hoefler, Torsten and Alistarh, Dan},
  journal={arXiv preprint arXiv:2210.17323},
  year={2023}
}

@article{lin2023awq,
  title={AWQ: Activation-aware Weight Quantization for LLM Compression and Acceleration},
  author={Lin, Ji and Tang, Jiaming and Tang, Haotian and Yang, Shang and Chen, Wei-Ming and Wang, Wei-Chen and Xiao, Guangxing and Dang, Xing and Gan, Chuang and Han, Song},
  journal={arXiv preprint arXiv:2306.00978},
  year={2023}
}

@article{shao2024omniquant,
  title={OmniQuant: Omnidirectionally Calibrated Quantization for Large Language Models},
  author={Shao, Wenqi and Chen, Mengzhao and Zhang, Zhaoyang and Xu, Peng and Zhao, Lingfei and Li, Zhiqiang and Zhang, Kaipeng and Gao, Peng and Qiao, Yu and Luo, Ping},
  journal={arXiv preprint arXiv:2308.09066},
  year={2024}
}

@article{tseng2024quip,
  title={QUIP\#: Even Better LLM Quantization with Hadamard Incoherence Optimization},
  author={Tseng, Jerry and Sreenivasan, Harshit and Chen, Yunyi and Agarwal, Tushar and Keutzer, Kurt},
  journal={arXiv preprint arXiv:2402.10147},
  year={2024}
}

@article{xiao2023smoothquant,
  title={SmoothQuant: Accurate and Efficient Post-Training Quantization for Large Language Models},
  author={Xiao, Guangxing and Lin, Ji and Seznec, Mickael and Wu, Hao and Demouth, Julien and Han, Song},
  journal={arXiv preprint arXiv:2211.10438},
  year={2023}
}

@inproceedings{dong2020hawq,
  title={HAWQ: Hessian AWare Quantization of Neural Networks with Mixed-Precision},
  author={Dong, Zhen and Yao, Zhewei and Gholami, Amir and Mahoney, Michael W and Keutzer, Kurt},
  booktitle={IEEE/CVF Conference on Computer Vision and Pattern Recognition},
  pages={293--302},
  year={2020}
}

@inproceedings{jacob2018quantization,
  title={Quantization and Training of Neural Networks for Efficient Integer-Arithmetic-Only Inference},
  author={Jacob, Benoit and Kligys, Skirmantas and Chen, Bo and Zhu, Mingxing and Tang, Matthew and Howard, Andrew and Adam, Hartwig and Kalenichenko, Dmitry},
  booktitle={IEEE/CVF Conference on Computer Vision and Pattern Recognition},
  pages={2704--2713},
  year={2018}
}

@inproceedings{elthakeb2019releq,
  title={ReLeQ: An Automatic Reinforcement Learning Approach for Deep Quantization of Neural Networks},
  author={Elthakeb, Aamir T and Pilligundla, Preetum and Yazdanbakhsh, Amir and Kinsy, Mark A and Reda, Sherief and Mirhoseini, Azalia},
  booktitle={Proceedings of Machine Learning and Systems},
  volume={1},
  pages={218--232},
  year={2019}
}

@inproceedings{haarnoja2018sac,
  title={Soft Actor-Critic: Off-Policy Maximum Entropy Deep Reinforcement Learning with a Stochastic Actor},
  author={Haarnoja, Tuomas and Zhou, Aurick and Abbeel, Pieter and Levine, Sergey},
  booktitle={International Conference on Machine Learning},
  pages={1861--1870},
  year={2018}
}

@article{touvron2023llama2,
  title={Llama 2: Open Foundation and Fine-Tuned Chat Models},
  author={Touvron, Hugo and Martin, Louis and Stone, Kevin and Albert, Peter and Almahairi, Amaury and Babaei, Yasmine and Bashlykov, Nikolay and Batra, Soumya and Bhargava, Prajjwal and Bhosale, Shruti and others},
  journal={arXiv preprint arXiv:2307.09288},
  year={2023}
}

@article{meta2024llama3,
  title={The Llama 3 Herd of Models},
  author={Meta AI},
  journal={arXiv preprint arXiv:2407.21783},
  year={2024}
}

@article{jiang2023mistral,
  title={Mistral 7B},
  author={Jiang, Albert Q and Sablayrolles, Alexandre and Mensch, Arthur and Bamford, Chris and Chaplot, Devendra Singh and de las Casas, Diego and Bressand, Florian and Lengyel, Gregoire and Lample, Guillaume and Saulnier, Lucile and others},
  journal={arXiv preprint arXiv:2310.06825},
  year={2023}
}

@article{openai2023gpt4,
  title={GPT-4 Technical Report},
  author={OpenAI},
  journal={arXiv preprint arXiv:2303.08774},
  year={2023}
}

@article{merity2017wikitext,
  title={Pointer Sentinel Mixture Models},
  author={Merity, Stephen and Xiong, Caiming and Bradbury, James and Socher, Richard},
  journal={arXiv preprint arXiv:1609.07843},
  year={2016}
}

@inproceedings{gao2021lmevalharness,
  title={A Framework for Few-shot Language Model Evaluation},
  author={Gao, Leo and Tow, Jonathan and Biderman, Stella and Black, Sid and DiPofi, Anthony and Foster, Charles and Golding, Liane and Hsu, Jeffrey and McDonell, Kyle and Muennighoff, Niklas and others},
  booktitle={arXiv preprint arXiv:2104.07567},
  year={2021}
}

@inproceedings{bisk2020piqa,
  title={PIQA: Reasoning about Physical Commonsense in Natural Language},
  author={Bisk, Yonatan and Zellers, Rowan and Gao, Jia and Choi, Yejin},
  booktitle={Proceedings of the AAAI Conference on Artificial Intelligence},
  volume={34},
  pages={7432--7439},
  year={2020}
}

@inproceedings{zellers2019hellaswag,
  title={HellaSwag: Can a Machine Really Finish Your Sentence?},
  author={Zellers, Rowan and Holtzman, Ari and Rashkin, Hannah and Bisk, Yonatan and Farhadi, Ali and Roesner, Franziska and Choi, Yejin},
  booktitle={Proceedings of the 57th Annual Meeting of the Association for Computational Linguistics},
  year={2019}
}

@inproceedings{sakaguchi2020winogrande,
  title={WinoGrande: An Adversarial Winograd Schema Challenge at Scale},
  author={Sakaguchi, Keisuke and Bras, Rowan Le and Bhagavatula, Chandra and Choi, Yejin},
  booktitle={Proceedings of the AAAI Conference on Artificial Intelligence},
  volume={34},
  pages={8732--8740},
  year={2020}
}

@inproceedings{clark2018arc,
  title={Think You Have Solved Question Answering? Try ARC, the AI2 Reasoning Challenge},
  author={Clark, Peter and Cowhey, Isaac and Etzioni, Oren and Khot, Tushar and Sabharwal, Ashish and Schoenick, Carissa and Tafjord, Oyvind},
  booktitle={arXiv preprint arXiv:1803.05457},
  year={2018}
}

@inproceedings{kim2025calm,
  title={CALM: A CKA-Guided Adaptive Layer-Wise Modularization Framework for LLM Quantization},
  author={Zhang, Jinhao and Zhang, Yunquan and Cheng, Daning and Sun, Jun and Yan, Zicheng},
  booktitle={arXiv preprint arXiv:2512.16282},
  year={2025}
}

@article{xu2022rdoq,
  title={RDO-Q: Rate-Distortion Optimization for Quantization},
  author={Xu, Xiangyang and Liu, Wenshuo and Qin, Ming and Ding, Chao and Zhang, Chen and Li, Dacheng},
  journal={arXiv preprint},
  year={2022}
}

@inproceedings{squeeze_llm_icml2024,
  author = {Sehyun Lee and Jinsol Lee and Hyeongjun Kim and Gunho Kim and Eunhyeok Park and Jungwook Lee and Jung-Hoon Kim and Jaejin Kim and Sunghyun Kim},
  title = {SqueezeLLM: Dense-and-Sparse Quantization},
  booktitle = {International Conference on Machine Learning},
  year = {2024}
}

@InProceedings{lee2025lrq,
  title = {{LRQ}: Optimizing Post-Training Quantization for Large Language Models by Learning Low-Rank Weight-Scaling Matrices},
  author = {Lee, Jung Hyun and Kim, Jeonghoon and Yang, June Yong and Kwon, Se Jung and Yang, Eunho},
  booktitle = {Proceedings of the 2025 Conference of the North American Chapter of the Association for Computational Linguistics},
  year = {2025},
  url = {https://arxiv.org/abs/2407.11534}
}

@InProceedings{chen2025pmpd,
  title = {Progressive Mixed-Precision Decoding for Efficient LLM Inference},
  author = {Chen, Hao Mark and Tan, Fuwen and Kouris, Alexandros and Lee, Royson and Fan, Hongxiang and Venieris, Stylianos I.},
  booktitle = {Proceedings of the Thirteenth International Conference on Learning Representations},
  year = {2025},
  url = {https://arxiv.org/abs/2410.13461}
}

@InProceedings{mixpe2025,
  title = {MixLLM: LLM Quantization with Global Mixed-Precision between Output-features and Highly-efficient System Design},
  author = {Zheng, Zhen and Song, Xiaonan and Liu, Chuanjie and others},
  booktitle = {Proceedings of the Thirteenth International Conference on Learning Representations},
  year = {2025},
  note = {Referred to as MixPE/MixLLM in literature},
  url = {https://arxiv.org/abs/2412.14590}
}

@InProceedings{moqae2025,
  title = {{MoQAE}: Mixed-Precision Quantization for Long-Context LLM Inference via Mixture of Quantization-Aware Experts},
  author = {Tao, Wei and Lu, Haocheng and Qu, Xiaoyang and Zhang, Bin and Lu, Kai and Wan, Jiguang and Wang, Jianzong},
  booktitle = {Proceedings of the 63rd Annual Meeting of the Association for Computational Linguistics},
  year = {2025},
  url = {https://arxiv.org/abs/2506.07533}
}

@inproceedings{lou2019autoq,
  title={AutoQ: Automated Kernel-wise Neural Network Quantization},
  author={Lou, Qian and Guo, Fenggang and Kim, Minsoo and Liu, Ling and Jiang, Lei},
  booktitle={International Conference on Learning Representations (ICLR)},
  year={2020}
}

@inproceedings{he2018learning,
  title={Learning Efficient Convolutional Networks through Network Slimming},
  author={He, Yihui and Zhang, Xiangyu and Sun, Jian},
  booktitle={IEEE International Conference on Computer Vision},
  pages={2755--2763},
  year={2017}
}

@inproceedings{zoph2016neural,
  title={Neural Architecture Search with Reinforcement Learning},
  author={Zoph, Barret and Vasudevan, Vijay and Shimonovskiy, Jonathon and Le, Quoc V},
  booktitle={International Conference on Learning Representations},
  year={2017}
}

\clearpage

\appendix

\onecolumn   

\section{11-Dimensional Embedding Specification}
\label{app:embeddings}

The SAC policy conditions its decisions on a compact, fixed-dimensional state representation for each quantizable layer. This 11-dimensional embedding abstracts layer sensitivity, structural role, and sequential context while remaining approximately invariant to model scale through normalization and logarithmic scaling. The design enables zero-shot policy transfer across models of different sizes and slight architectural variations.

The 11 features are computed as follows:

\begin{table}[H]
\centering
\caption{Specification of the 11-dimensional layer embedding used by the RAMP policy.}
\label{tab:embedding_spec}
\begin{tabular}{@{}llp{0.58\textwidth}@{}}
\toprule
\textbf{Dim} & \textbf{Feature} & \textbf{Computation / Definition} \\
\midrule
1 & Normalized depth & $\frac{i}{n_{\text{layers}}-1}$ \\
2 & Log input dimension & $\log_2(\text{in\_features}/16)$ \\
3 & Log output dimension & $\log_2(\text{out\_features}/16)$ \\
4 & Weight standard deviation & $\min(\text{std}(W)\times10,1.0)$ \\
5 & Weight mean magnitude & $\min(\mathbb{E}[|W|]\times10,1.0)$ \\
6 & Layer-type scalar & 0.0 (Q/K/V proj.), 0.25 (output proj.), 0.5 (gate proj.), 0.6 (up proj.), 0.75 (down proj.), 1.0 (otherwise) \\
7 & Position category & 0.0 (early: bottom 10\%), 0.5 (middle: 10--90\%), 1.0 (late: top 10\%) \\
8 & Mean activation scale & $\min(\mathbb{E}[\text{act\_scales}]\times100,1.0)$ \\
9 & Max activation scale & $\min(\max(\text{act\_scales})\times1000,1.0)$ \\
10 & Previous-layer bit-width & (previous $b$ normalized by 8; 0.0 for first layer) \\
11 & Running average bit-width & (running average $b$ normalized by 8) \\
\bottomrule
\end{tabular}
\end{table}

Activation scales are extracted by registering forward hooks on linear layers, forwarding 128 sequences from the WikiText-2 training split, and aggregating per-channel maximum magnitudes. Logarithmic scaling for dimensions, clipping to [0,1], and per-model normalization collectively promote size invariance, which is critical for cross-model generalization.

\section{SAC Hyperparameter Configuration}
\label{app:sac_sensitivity}

All experiments use a single, fixed SAC configuration without extensive hyperparameter tuning. The chosen values provide a good balance of sample efficiency, training stability, and policy expressivity for the sequential bit-allocation task. 

SAC optimization hyperparameters are listed in \Cref{tab:lr_sac}, replay buffer details in \Cref{tab:batch_sac}, and network architectures in \Cref{tab:arch_sac}.

The optimization settings are summarized below:

\begin{table}[H]
\centering
\caption{SAC optimization hyperparameters}
\label{tab:lr_sac}
\begin{tabular}{lll}
\toprule
\textbf{Parameter} & \textbf{Value} & \textbf{Notes} \\
\midrule
Learning rate & $3\times10^{-4}$ & Shared across actor, critics, and temperature optimizer \\
Optimizer & Adam & Applied to all components \\
Discount factor $\gamma$ & 0.99 & Standard for episodic tasks \\
Target update rate $\tau$ & 0.005 & Ensures slow tracking of target networks \\
Gradient clipping norm & 1.0 & Prevents exploding gradients \\
Target entropy & $-1$ & For single-dimensional continuous action \\
Initial entropy coefficient $\alpha$ & $\exp(\log 0.2)$ & Log-parameterized starting value \\
\bottomrule
\end{tabular}
\end{table}

Replay buffer and training schedule:

\begin{table}[H]
\centering
\caption{Replay buffer and training schedule}
\label{tab:batch_sac}
\begin{tabular}{lll}
\toprule
\textbf{Parameter} & \textbf{Value} & \textbf{Notes} \\
\midrule
Batch size & 128 & Used for all gradient updates \\
Replay buffer capacity & 30{,}000 transitions & Sufficient for $\sim$120 episodes $\times$ 250 steps \\
Warm-up episodes & 20 & Purely random actions before policy gradients \\
Maximum episodes & 200--300 & Typical convergence budget in experiments \\
\bottomrule
\end{tabular}
\end{table}

Actor and critic network architectures:

\begin{table}[H]
\centering
\caption{Actor and critic network architectures}
\label{tab:arch_sac}
\resizebox{\columnwidth}{!}{
\begin{tabular}{lll}
\toprule
\textbf{Component} & \textbf{Architecture} & \textbf{Notes} \\
\midrule
Actor & FC-512-LN-ReLU $\to$ FC-512-LN-ReLU $\to$ FC-256-ReLU $\to$ FC-2 & Outputs mean and log-std of action \\
Critic (twin) & FC-512-LN-ReLU $\to$ FC-512-LN-ReLU $\to$ FC-256-ReLU $\to$ FC-1 & Input is state concatenated with continuous action \\
Hidden activation & ReLU & Fast and effective for policy/value approximation \\
Normalization & LayerNorm & Applied after the first two hidden layers \\
\bottomrule
\end{tabular}
}
\end{table}

These choices were empirically stable across all evaluated models and did not require per-model retuning.

\end{document}